\documentclass[11pt]{article}
\usepackage[letterpaper,margin=1in]{geometry}
\usepackage{microtype}
\usepackage{graphicx}
\usepackage{subcaption}
\usepackage{booktabs}
\usepackage{longtable}
\usepackage{times}
\usepackage[numbers,sort&compress]{natbib}
\usepackage[hidelinks]{hyperref}
\usepackage{url}
\usepackage{amsmath,amssymb,amsfonts}
\usepackage{multirow}
\usepackage{xcolor}
\usepackage{pifont}
\usepackage{flafter}
\usepackage{algorithm}
\usepackage{algorithmic}
\usepackage{tikz}
\usetikzlibrary{arrows.meta,positioning,fit,backgrounds,calc}
\usepackage{fancyhdr}

\newcommand{\zL}{z_{L}}
\newcommand{\zH}{z_{H}}
\newcommand{\zLc}{z_{L}^{\mathrm{c}}}
\newcommand{\zA}{z_{\mathrm{ACT}}}
\newcommand{\dz}{\Delta z}

\providecommand{\tabwidth}{\columnwidth}

\providecommand{\papertitle}{When Wrong Answers Remain Unsettled:\\Failure Dynamics in Recurrent Reasoners}
\newenvironment{audittable}[1]{\begin{center}\begin{tabular}{#1}}{\end{tabular}\end{center}}
\newif\ificlr
\iclrfalse

\providecommand{\panelaw}{0.40}
\providecommand{\panelbw}{0.37}

\providecommand{\fittab}[1]{\resizebox{\tabwidth}{!}{#1}}

\providecommand{\figw}{\linewidth}

\renewenvironment{audittable}[1]{\begin{longtable}{#1}}{\end{longtable}}
\iclrtrue   
\newlength{\icmlcol}
\renewcommand{\tabwidth}{\icmlcol}
\renewcommand{\panelaw}{0.4909}
\renewcommand{\panelbw}{0.4541}
\renewcommand{\fittab}[1]{#1}
\renewcommand{\figw}{0.82\linewidth}

\title{\papertitle}
\author{%
  Jiucheng Zang \quad Congkai Peng \quad Yankai Wei\\[2pt]
  University of Waterloo (uwaterloo.ca)%
}
\date{}

\begin{document}
\maketitle
\begin{abstract}
When a recurrent reasoner returns a wrong answer, is its computation finished
or merely stopped? We test this on a trained Hierarchical Reasoning Model
(HRM; 27M parameters, Sudoku-Extreme) by running the recurrence for one more
inference budget and comparing with an unperturbed control.
Solved answers persist ($0.95$ of $512$ endpoints). Wrong answers mostly do
not: $0.42$ of near-miss answers ($n=170$) and $0.18$ of all $512$ wrong
endpoints keep their answer. Perturbations from $0.05\times$ to $20\times$ the
terminal-update scale barely change these rates, so the measurement reflects
continued recurrence, not noise sensitivity.
The solved-above-wrong ordering also holds in five further settings (a second
seed, TRM, Maze, GRAM-guided TRM, N-Queens), but a fixed $\tfrac12$ persistence
threshold does not, and neither do HRM's endpoint-spectrum and fast-versus-slow
predictability patterns, which are not universal diagnostics.
Perturb-and-reselect (known from prior work) is consistent with this: frozen
noise gains $4.8$ accuracy points ($197$ gained, $2$ lost), but compute-matched
extended depth gains $5.6$ and Sudoku symmetry augmentation $34.6$, so noise
is not the best use of the budget.
Conclusions are limited to finite horizons and one primary HRM checkpoint;
they do not exclude asymptotic attractors.

\end{abstract}

\section{Introduction}
\label{sec:intro}

Recurrent-depth models increase inference computation by iterating weight-shared
blocks, from Universal Transformers \citep{ut2018} to HRM \citep{hrm2025} and
latent recurrent-depth models \citep{geiping2025,saunshi2025latent}.
Unlike a standard transformer's sequence of distinct operators
\citep{logitlens2020,tunedlens2023}, a tied map $z_{t+1}=f(z_t,x)$ defines a
dynamical system that can be analysed directly \citep{sussillo2013}.
This provides a way to study how test-time compute \citep{snell2024ttc}
changes a prediction, including whether further iteration approaches a fixed
point \citep{bai2019deq}.

When such a model returns a wrong answer, is its computation finished, or has
it merely stopped? Recent studies describe failure as attractor entrapment,
bounded non-convergence, continued exploration, or a first fixed point that
happens to be wrong
\citep{efstathiou2026,ptrm2026,corbett2026,renliu2026}. For TRM, extended
recurrence already shows that many unsolved puzzles are solved later while a
persistent subset is not \citep{krishna2026}. What is missing for HRM is a
population-level measurement that asks the question directly: does the
terminal answer survive further recurrence, and does an unperturbed control
separate continuation from noise? A flat decoded-loss curve or a clustered PCA
projection does not answer it (Sec.~\ref{sec:related}).

\paragraph{Takeaway.} In the primary HRM checkpoint, wrong endpoints are
mostly \emph{unsettled}: after one more inference budget, $0.95$ of solved
answers persist but only $0.18$ of wrong ones ($0.42$ for near-misses). Prior
qualitative accounts of ``entrapment'' therefore do not describe most HRM
failures at this horizon. The geometric diagnostics we also report
(endpoint spectra, fast-versus-slow predictability) are specific to HRM and do
not transfer to TRM or GRAM-guided TRM, so they should not be used as general
failure explanations.

We measure this persistence directly, comparing perturbed continuation with
an unperturbed control (Sec.~\ref{sec:fail}). In the primary HRM checkpoint,
wrong terminal answers are substantially less persistent than correct ones.
Endpoint Jacobian--vector products characterise the local linearisation
(Sec.~\ref{sec:endpoint}); progressively richer linear models isolate which
transient updates remain poorly predicted (Sec.~\ref{sec:transient}).
We relate these updates to constraint propagation
(Sec.~\ref{sec:compute}) and evaluate perturbation and reselection against
compute-matched alternatives (Sec.~\ref{sec:intervention}).

\paragraph{Contributions, and what is replication.}
We separate what is new from what prior work already established.
Perturb-and-reselect \citep{efstathiou2026,ptrm2026,corbett2026,illouz2026},
HRM's tendency to settle on a wrong fixed point \citep{renliu2026}, and
extended-recurrence gains on TRM \citep{krishna2026} are \emph{prior work};
our runs replicate them and add paired controls. The items below are what we
add.
\begin{enumerate}
  \item \textbf{Answer persistence with an unperturbed control.}
  After one additional inference budget, mild wrong answers persist at
  $0.42$ versus $0.95$ for solved endpoints ($n{=}170$ and $512$;
  $0.18$ pooled over all $512$ wrong endpoints).
  Perturbations up to $20\times$ the update scale have little effect on
  these rates. The measurement therefore primarily reflects continuation.
  The solved--wrong ordering holds across six evaluated settings, although
  a majority-persistence threshold does not
  (Fig.~\ref{fig:coverage-summary}; Apps.~\ref{app:eps0},~\ref{app:robust}).
  \item \textbf{Local and transient dynamics, with transfer limits.}
  Solved HRM endpoints have a mean leading-Ritz radius of $0.44$.
  Scheduled linear models predict fast-loop updates better than slow-loop
  updates (held-out $R^2{=}0.72$ versus $0.37$).
  These geometric patterns change in TRM and GRAM-guided TRM, limiting
  their use as general explanations of failure
  (Secs.~\ref{sec:endpoint}--\ref{sec:transient}).
\end{enumerate}
Our reselection evaluation adds paired gains and regressions, injection-site comparisons,
oracle headroom, and compute-matched baselines: noise injection gains
$4.8$ points ($197$ gained / $2$ lost), while extended depth performs better.
We also document mean-field pooling and incomplete clock conditioning as
sources of misleading dynamical summaries, with synthetic regression tests
(App.~\ref{app:synthetic}).

\begin{figure*}[t]
\centering
\begin{minipage}[c]{\panelaw\linewidth}\centering
\resizebox{\linewidth}{!}{\begin{tikzpicture}[font=\small,>={Stealth[round]},
  box/.style={draw,rounded corners,align=center,minimum height=8mm,minimum width=15mm,inner sep=3pt},
  loop/.style={draw,densely dashed,rounded corners,gray}]
\node[box,fill=blue!8]   (L) {L-module\\ $\zL$\\[-2pt] {\tiny\color{blue!55!black}linear $R^2\,0.72$}};
\node[box,fill=orange!12,right=13mm of L] (H) {H-module\\ $\zH$\\[-2pt] {\tiny\color{orange!62!black}linear $R^2\,\mathbf{0.37}$}};
\begin{scope}[on background layer]
\node[loop,fit=(L),inner sep=4pt] (l1){};
\node[loop,fit=(L)(l1),inner sep=7pt] (l2){};
\node[loop,fit=(L)(H)(l2),inner sep=10pt] (l3){};
\node[draw=orange!75!black,line width=0.7pt,rounded corners,fill=orange!35,fit=(H),inner sep=3pt] (slow){};
\end{scope}
\draw[->,orange!75!black,line width=0.6pt]
  (slow.south east) to[out=-35,in=-145,looseness=0.8] (slow.south west);
\node[left=7mm of L] (x) {$x$};
\draw[->] (x) -- (L);
\draw[->] (L) -- node[above,font=\scriptsize]{$\zL$} (H);
\draw[->,densely dashed] (H.north) to[out=110,in=70] node[above,font=\scriptsize]{$\zH{+}x$} (L.north);
\node[right=6mm of H,align=left,font=\scriptsize] (o) {q-halt,\\ lens};
\draw[->] (slow.east) -- (o);
\node[font=\scriptsize,gray,anchor=north,inner sep=0pt] (clk) at ($(l3.south)+(0,-1.6mm)$)
  {segment $s\!\supset\!$ H-cycle $h\!\supset\!$ L-step $\ell$};
\node[font=\scriptsize,anchor=north,inner sep=0pt,align=left] (sit) at ($(clk.south)+(0,-2.4mm)$)
  {\begin{tabular}{@{}l@{\ }r@{\,}l@{\ }l@{}}
   \multicolumn{4}{@{}l}{\emph{three injection sites; over $41.6\%$:}}\\
   inject $x$ & $-1.1$ & pts, & \textbf{243} lost\\
   \color{blue!55!black}inject $\zL$ & $+2.9$ & pts, & 1 lost\\
   \color{orange!62!black}inject $\zH$ & $+4.6$ & pts, & 2 lost\\[1pt]
   \multicolumn{4}{@{}l}{\emph{no injection, same $9\times$ budget:}}\\
   depth alone & $+5.6$ & pts, & 14 lost\\
   depth, best seg. & $\mathbf{+6.0}$ & pts, & 2 lost $\leftarrow$ beats all three\\
   \end{tabular}};
\end{tikzpicture}}\\[2pt]
{\scriptsize (a) mechanism $\to$ site $\to$ gain}
\end{minipage}\hspace{7mm}%
\begin{minipage}[c]{\panelbw\linewidth}\centering
\includegraphics[width=\linewidth]{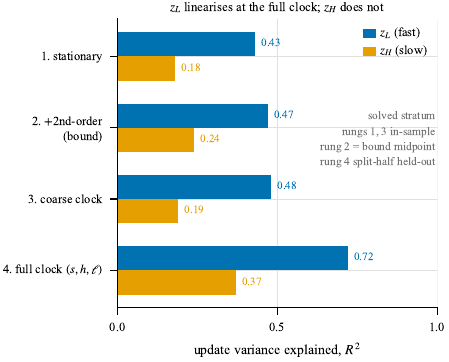}\\[2pt]
{\scriptsize (b) progressive exclusion of linear descriptions}
\end{minipage}
\caption{\textbf{Overview.} \textbf{(a)} HRM iterates two weight-tied modules on
an exogenous three-level clock, the input injection $\zH{+}x$ entering the fast
L-module each step; shaded is the slow $\zH$ loop, which (b) isolates. Site
gains: one $k{=}8$ run at matched radii, $N{=}4096$ (App.~\ref{app:ablation}).
Recurrent states are perturbed after five segments; input is perturbed before
recurrence and remains perturbed throughout. Depth is evaluated in
App.~\ref{app:depth}; Sec.~\ref{sec:intervention}'s headline $+4.8$ is
this same $\zH$ site under the frozen recipe. The site axis identifies
\emph{why} the input site fails, not the best use of a test-time budget.
\textbf{(b)} Excluding successively stronger linear descriptions of the update
trajectory leaves the slow loop least captured: held-out $R^2{=}0.37$ for $\zH$
against $0.72$ for the fast $\zL$, a close-to-scheduled linear program.}
\label{fig:overview}
\end{figure*}
\section{Related Work}
\label{sec:related}

\paragraph{Recurrent-depth models and their dynamics.}
HRM \citep{hrm2025} and TRM \citep{trm2025} reuse parameters across
recurrent updates, building on universal transformers \citep{ut2018}
and adaptive computation time \citep{graves2016act}, with related
equilibrium formulations in deep equilibrium models \citep{bai2019deq}.
Recent work scales latent depth \citep{geiping2025} and asks how looped
models use it
\citep{loopedmech2026,depthuse2025,twoscale2025,grow2loop2026,dram2026}.
\citet{renliu2026} study HRM's trajectory geometry and input sensitivity;
we examine endpoint spectra and whether a terminal state keeps its answer
when continued or perturbed. The input-perturbation effect they report
independently is what our input-site injection probes, separately from
injection into the recurrent state (App.~\ref{app:ablation}).
Fixed-point linearisation \citep{sussillo2013} has limited reach away
from equilibrium \citep{linearization2023}, motivating trajectory-level
analyses \citep{dsa2023,lineattractor2019}. Whereas Jacobian-switching
LDS \citep{jslds2021} uses state-dependent linearisation, our full-clock
test fits operators indexed by the update schedule, capturing HRM's fast
$\zL$ updates better than its slow $\zH$ updates
(Sec.~\ref{sec:transient}).

\paragraph{Perturbation and reselection.}
Sample-and-select methods \citep{wang2022selfconsistency} require diverse
candidates. For deterministic HRM, these can come from task symmetries
(App.~\ref{app:tta}) or noise; GRAM \citep{gram2026} incorporates
stochasticity into the recurrence. Prior methods perturb the answer initialisation and scale the
number of attempts \citep{efstathiou2026}, inject noise
throughout recursion \citep{ptrm2026}, or use online particle filtering
\citep{corbett2026}, with learned halting scores guiding selection.
Our independent reruns use a single injection and terminal halting-head
selection. We add paired gains and regressions, site comparisons and
compute-matched alternatives, alongside an exploratory timing analysis
(App.~\ref{app:timing}). Candidate diversity and guide quality are also
studied in prior work; we do not claim to introduce that distinction.

\paragraph{Descriptions of failure and terminal-answer persistence.}
Prior work discusses attractor entrapment \citep{efstathiou2026}, bounded
non-convergence \citep{ptrm2026}, and continued exploration
\citep{corbett2026}. \citet{efstathiou2026} state that on Sudoku-Extreme
``failed runs plateau at stable high-loss attractors rather than continuing
exploration'', which is the description our measurement addresses. We quote
their abstract; their full text was unavailable, so we do not describe the
experiments behind it (App.~\ref{app:disagreement}).
\citet{renliu2026} report that HRM can settle on a wrong fixed point, and
\citet{krishna2026} quantify answer changes under extended recurrence in TRM,
with a Jacobian analysis that reports a trajectory-aligned gain and a
worst-case singular value. We replicate the qualitative picture on HRM with a
solved-versus-wrong contrast and an unperturbed control; our endpoint
spectral radius is a different, direction-agnostic quantity, and neither
replaces their measurement. \citet{illouz2026} find that resampling the decoded
answer state recovers more failures than perturbing the latent state; our
injection-site comparison (App.~\ref{app:ablation}) is a separate HRM test
and does not reproduce that setting. \citet{chen2026ebr} take as a premise
that convergence does not imply correctness.
\citet{corbett2026} provide tube-stability diagnostics under a contraction
assumption. Our endpoint spectra address a different quantity: a local
leading-eigenvalue estimate, rather than a uniform contraction bound along a
trajectory. The two cannot be substituted for one another.
We test whether the terminal decoded answer persists beyond the trained
budget, using an unperturbed control to distinguish continuation from noise
effects. Finite continuation does not establish the absence of asymptotic
attractors (App.~\ref{app:horizon}).
\section{Background and Experimental Setup}
\label{sec:setup}

\paragraph{Model and task.}
The Hierarchical Reasoning Model (HRM) is a $27.3$M-parameter recurrent
transformer of two coupled recurrences (a fast L-module and a slow H-module) wrapped
in an adaptive-computation-time (ACT) outer loop with a learned q-halt head. Blocks
are post-norm with RMSNorm; the recurrences couple by additive input injection, $\zH{+}x$ into the L-module each L-step and $\zL$ into the H-module once
per H-cycle. We study the released configuration $H{=}2$, $L{=}2$ with up to $16$ ACT
segments, trained on Sudoku-Extreme ($1000$ puzzles $\times$ $1000$ augmentations;
vocab $11$, sequence length $81$) for $20{,}000$ epochs on a single RTX~6000 Ada
($10$h$44$m; global batch $384$, lr $7\mathrm{e}{-}5$; no pretraining, no
chain-of-thought). An epoch is one pass over the $1000$ base puzzles, not over
the augmented million: $52{,}080$ steps at batch $384$ is $1000$ examples per
epoch. We analyse checkpoint \texttt{step\_52080}
(epoch $20{,}000$), which attains exact accuracy $41.76\%$ (per-cell $79.92\%$,
q-halt accuracy $99.46\%$) on the full $422{,}786$-puzzle test set.
The primary run uses seed $0$, \texttt{lr\_min\_ratio=1.0} and
\texttt{weight\_decay=1.0}; the remaining configuration is recorded with the checkpoint.

\begin{figure*}[t]
\centering
\includegraphics[width=\linewidth]{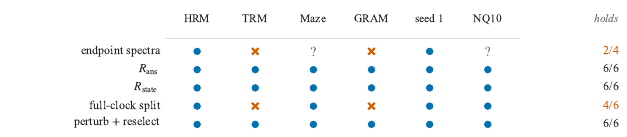}
\caption{\textbf{What travels.} Each claim against the settings that can carry
it---disc holds, cross fails, query undecided---tallied over the settings that
returned a verdict. The two that break the geometry are the two that share a
collapsed module. Values, $n$ and caveats: Fig.~\ref{fig:coverage}.}
\label{fig:coverage-summary}
\end{figure*}

\paragraph{What was run.} Body measurements are HRM/Sudoku-Extreme, seed-0. The
same diagnostics run unchanged on five further settings---a second seed, a second
architecture (TRM), a second task (Maze-$30{\times}30$-Hard), a GRAM-guided
recurrence, and a third task (N-Queens-$10{\times}10$).
Fig.~\ref{fig:coverage-summary} is which of the five claims each one carries. A
sixth, N-Queens-$12{\times}12$, did not train and is reported as a null
(App.~\ref{app:nqueens}). The intervention is reported against compute-matched
alternatives---depth alone (App.~\ref{app:depth}) and symmetry augmentation
(App.~\ref{app:tta})---and the stationarity rates against two controls: an
unperturbed rerun (App.~\ref{app:eps0}) and continuation to $128\times$ the
trained budget (App.~\ref{app:horizon}). Thirteen of the $63$ job scripts behind these
carry an interpretation rule committed before execution (App.~\ref{app:prereg}).

\paragraph{Three timescales and notation.}
An exogenous clock nests the computation into three levels (ACT segment
$s\!\in\!\{1..16\}$, H-cycle $h$, L-step $\ell$, with $H{=}L{=}2$;
Fig.~\ref{fig:overview}(a)), driving two coupled
updates,
\begin{equation}
\zL \leftarrow \mathrm{L}(\zL,\, \zH{+}x), \qquad \zH \leftarrow \mathrm{H}(\zH,\, \zL).
\label{eq:recurrence}
\end{equation}
We study the
residual-stream updates $\dz_t{=}z_{t+1}{-}z_t$ at each level's timescale ($\zL$ within
an H-cycle, $\zH$ across H-cycles, $\zA$ across segments). Hidden states are
$512$-dimensional at each of $81$ Sudoku cells (plus a puzzle embedding); every
statistic treats each $(t,\text{position})$ update as one sample.

\paragraph{Reduction discipline (read first).}
All trajectory statistics use \emph{per-position} resolution;
position-averaged (``mean-field'') reductions inflate linearity and complex mass and are shown only where flagged
(see App.~\ref{app:synthetic}).

\paragraph{Evaluation samples.}
Unless noted, analyses draw on a stratified sample of $512$ solved and $512$
unsolved puzzles, the unsolved bucketed by wrong-cell count ($1$--$5$ / $6$--$20$
/ ${>}20$); the recomputed solved rate matches the official $41.76\%$ to within
$0.1$pt. Intervention deltas use a disjoint uniform $4{,}096$-puzzle subset.
Every $\Delta$ in points is paired on those same puzzles:
$\Delta{=}100(b{-}c)/N$ for $b$ gained (wrong$\to$right) and $c$ lost, on
\emph{exact} (whole-grid) accuracy; intervals use a paired normal approximation,
$\mathrm{Var}(\Delta){=}10^4(b{+}c{-}(b{-}c)^2/N)/N^2$, while regression
\emph{rates} use Wilson.

\paragraph{Two senses of ``settled''.} The \emph{terminal-update gate} asks
whether the last step is small: $\|\dz_{\mathrm{final}}\|/\|z\|\le q_{95}$, for
$q_{95}$ the $95$th percentile of that \emph{relative} ratio over solved
endpoints. Post-norm RMSNorm pins $\|z\|$, so the denominator varies little and
the cut is close to an absolute one on $\|\dz_{\mathrm{final}}\|$
(App.~\ref{app:stepsize}). \emph{Stationary} asks whether continued computation leaves the state
where it is (Eq.~\ref{eq:stationarity}). They are not the same question: $95\%$
of solved but also $69\%$ of wrong endpoints clear the gate
(App.~\ref{app:jacobian}), and Sec.~\ref{sec:fail} is about that gap. We call these
\emph{gate-passing} states; the data-calibrated gate does not certify a fixed point.
\section{Terminal-answer persistence and correctness prediction}
\label{sec:fail}

\textbf{Wrong terminal answers are less persistent than correct ones.}
We compare decoded-answer persistence, $R_{\mathrm{ans}}$, with state
persistence, $R_{\mathrm{state}}$. The latter uses a setting-specific distance
scale and is not directly comparable across models
(Sec.~\ref{sec:discussion}). Reselection is evaluated separately in
Sec.~\ref{sec:intervention}.

\paragraph{One stationarity test.} Let $z_i^\ast$ be the terminal $\zH$ of
endpoint $i$, $\Phi$ the recurrence of Eq.~\ref{eq:recurrence} run for a further
$16$ segments (the trained budget, fixed for every endpoint rather than re-halted
by ACT), $\hat y(\cdot)$ the decoded answer on the answer mask, $S$ the
number of state entries ($82$ positions $\times\ 512$ channels $=41{,}984$), and
$\kappa$ one \emph{global} scale, the median
$\|\dz_{\mathrm{final}}\|$ over the analysed endpoints. The kick enters $\zH$
alone and $\zL$ passes into the continuation unchanged, so $\varepsilon{=}0$
reproduces the continuation the model itself would run. \emph{Perturb and
continue:}
\begin{equation}
z_{i,\varepsilon} = \Phi\big(z_i^\ast + \varepsilon\kappa S^{-1/2}\xi\big),
\quad \xi\sim\mathcal{N}(0,I_S),
\quad \mathbb{E}\big\|\varepsilon\kappa S^{-1/2}\xi\big\|^2=\varepsilon^2\kappa^2.
\label{eq:kick}
\end{equation}
The factor $S^{-1/2}$ makes the root-mean-square perturbation norm
$\varepsilon\kappa$, independently of state dimension.

\emph{Score the answer, then the class:}
\begin{equation}
r_i(\varepsilon) = \Pr\nolimits_\xi\big[\hat y(z_{i,\varepsilon}){=}\hat y(z_i^\ast)\big],
\qquad
R_{\mathrm{ans}}^{(C)}(\varepsilon)=\tfrac{1}{|C|}\textstyle\sum_{i\in C}r_i(\varepsilon).
\label{eq:stationarity}
\end{equation}
\textbf{Every rate we report is the class-level $R^{(C)}$}; $R_{\mathrm{state}}$
is the same with $\|z_{i,\varepsilon}{-}z_i^\ast\|{<}\kappa$ replacing the answer
test. \emph{Now set $\varepsilon{=}0$:}
\begin{equation}
R_{\mathrm{ans}}^{(C)}(0) = \tfrac{1}{|C|}\big|\{\,i\in C:\ \hat y(\Phi(z_i^\ast)){=}\hat y(z_i^\ast)\,\}\big|.
\label{eq:eps0}
\end{equation}
For the deterministic recurrence, $r_i(0)$ is binary and the class rate is
the fraction retaining its answer after one additional budget.
We use $R^{(C)}_{\mathrm{ans}}>\tfrac12$ as a descriptive
\emph{majority-persistence} criterion. It is a population summary, not an
attractor definition. Equality at the two readouts also does not exclude
intermediate answer changes.

The terminal-update gate and answer persistence address different questions:
$69\%$ of wrong HRM endpoints pass the former, yet most fail the latter.
Neither test alone certifies asymptotic stability.

\paragraph{Failure taxonomy.} Gate failure rises with error severity but stays a
\emph{minority} in every bucket---$21\%$ of mild failures, $29\%$ at $6$--$20$,
$43\%$ beyond (App.~\ref{app:jacobian})---so the
dominant failure mode is \emph{small-step-but-wrong}.

\paragraph{Persistence and replication.}
Solved endpoints retain their decoded answer at
$R_{\mathrm{ans}}=0.95\,[0.93,0.97]$ and remain within the state-distance
threshold at $R_{\mathrm{state}}=0.86\,[0.82,0.88]$.
For mild wrong endpoints these rates are $0.42\,[0.35,0.49]$ and $0.13$
(endpoint-clustered bootstraps; $n=512$ solved and $170$ mild wrong).
Mild is the most persistent wrong bucket under both dump and replay labels.
The other dump-labelled buckets score $0.09$ and $0.04$, but their ordering
reverses under replay labels (App.~\ref{app:relabel}).
The pooled rate of $0.18$ refers to the stratified sample.

Across $13$ radii in $[0.05,20]$ and $12$ seeds, persistence changes little;
the median solved-state displacement changes by $0.8\%$ between
$\varepsilon=0$ and $20$ in the pilot (App.~\ref{app:eps0}).
In longer continuation, wrong answers persist across equal $128$-segment
windows at a mean rate of $0.43$, versus $0.97$ for solved endpoints,
through $2048$ segments (App.~\ref{app:horizon}).
This measures agreement between readouts, not progress toward correctness.

The solved--wrong ordering also appears in TRM ($0.96/0.32$ answer
persistence; $0.32/0.02$ state persistence) and Maze.
Absolute state rates vary substantially across settings.
The answer-level $\tfrac12$ criterion separates the four Sudoku- and
Maze-based settings at every tested radius, but fails on N-Queens-$10\times10$
($0.96/0.56$) and GRAM-guided TRM ($0.98/0.84$).
Thus the observed ordering transfers across six settings; the fixed cut does
not (Fig.~\ref{fig:coverage}). Both also reproduce the control: unperturbed
continuation returns $0.59$ for N-Queens' wrong class against $0.56$ perturbed,
and $0.84$ for the GRAM-guided recurrence against $0.83$ at the control's own
matched radius, so what the rates
measure there is persistence and not perturbation sensitivity, as on HRM.
Severity mixtures still differ between settings, so the replications remain a
qualitative ordering.

\paragraph{What this does and does not show.}
Higher persistence for solved answers is expected: training supervises the
decoded answer at every segment, so correct answers are plausibly stable by
construction. The informative quantity is the wrong-class rate, which tests
accounts of failure as entrapment; at this horizon most wrong endpoints change
their answer. Two caveats apply. First, solved and wrong puzzles differ in
difficulty, and we did not run a difficulty-matched pairing; error-severity
buckets are only a partial control. Second, a changed answer is not a
corrected one, and persistence says nothing about why a puzzle fails.
The result concerns one primary checkpoint plus the replications above.

\paragraph{Does the model know? Probes vs.\ q-halt.}
A linear probe trained on the final $\zH$ (group 5-fold CV) reads correctness
at AUROC $0.941$ overall and $0.887$ on the mild small-step-but-wrong failures, barely
above the model's own halting head ($0.934$ and $0.873$ on the same splits; that
margin shrinks to zero under replay labels, App.~\ref{app:relabel}). The tested linear probe therefore adds little to q-halt on these samples.
This does not establish optimality over all linear readouts or absence of
correctness information: trajectory features and nonlinear readouts improve
discrimination (App.~\ref{app:nonlinear-readout}).

\begin{figure*}[t]
\centering
\includegraphics[width=\linewidth]{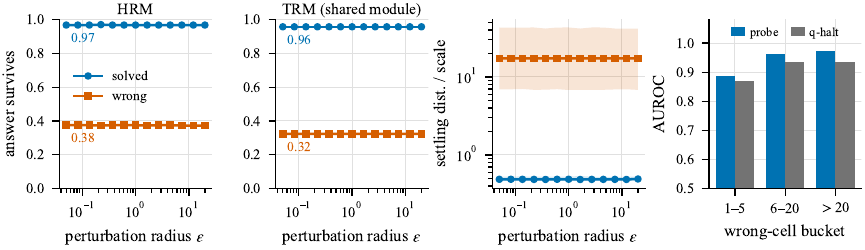}
\caption{Failure structure (solved vs.\ wrong). \textbf{(a,\,b) The central
criterion:} $R_{\mathrm{ans}}$, the fraction of continuations whose decoded
answer survives, against the radius $\varepsilon$ of an optional perturbation
($13$ radii $\times$ $12$ seeds $\times$ $32$ endpoints/class; $96$ for TRM).
\emph{The lines are flat because $\varepsilon$ does not matter}: the axis starts
at $\varepsilon{=}0.05$, and the separately-run unperturbed control returns these
same rates ($0.97/0.38$; App.~\ref{app:eps0}), so what is measured is whether the
terminal decoded answer settles under continued recurrence. TRM preserves the \emph{ordering}, not the magnitude.
The interval-bearing rates in the text come from the $512$-endpoint replication
(App.~\ref{app:eps0}), which trades radii for $n$. \textbf{(c)} State-level graded
form (median, IQR). \textbf{(d)} A correctness probe on $\zH$ barely beats the
model's own halting head.}
\label{fig:fail}
\end{figure*}
\section{Local linearisation at terminal states}
\label{sec:endpoint}

\textbf{Method.} We linearise the per-iteration H- and L-updates about solved
endpoints $z^\ast$ and wrong terminal states, which need not be exact roots.
Rather than materialise the $(SD)\times(SD)$ Jacobian $\partial f/\partial z$ we
probe it with Jacobian--vector products and $m{=}40$ Arnoldi steps, checked
against a dense eigendecomposition, and summarise each spectrum by
$\hat\rho_{40}{=}\max_i|\theta_i|$ over the retained Ritz values and their complex
fraction ($256$ endpoints per class, both maps $f_H,f_L$). Even at solved
endpoints, $\hat\rho_{40}<1$ does not rule out expanding modes missed by the
Krylov subspace and is not a stability certificate. The terminal-update gate
provides residual information, not a bound on unobserved modes.

\textbf{Results} (App.~Fig.~\ref{fig:jacobian_ecdf}; per-example complex-plane spectra
in App.~\ref{app:jacobian}): solved $f_H$ has low estimated radii
(mean $\rho{=}0.44$, every gate-passing endpoint below $0.90$,
$65\%$ of retained Ritz values complex; $f_L$ mean $\rho{=}0.78$), while
wrong $f_H$ has mean $\hat\rho_{40}{=}0.75$ (median $0.77$) with $2.7\%$ showing an
estimated expanding leading mode ($\hat\rho_{40}{>}1$; $7/256$, $95\%$ CI $[1.3,5.5]\%$) and $9.4\%$ above
$0.9$.
The tail is \emph{not} exclusive to failures: raw it points both ways, and gated
it is one-sided, $0/243$ solved vs.\ $4/176$ wrong (App.~\ref{app:jacobian}). The
summary we rely on is neither tail but the distributional one:
$\mathrm{AUC}{=}0.913$, $95\%$ bootstrap CI $[0.886,0.940]$, unchanged under replay
labels ($0.928$; App.~\ref{app:relabel}).

At solved endpoints with small final updates, these estimates are consistent
with locally damped dynamics containing oscillatory modes.
They do not certify an exact fixed point, Euclidean contraction, or stability
of the coupled recurrence. We next examine the transient preceding these
endpoints (App.~Fig.~\ref{fig:trajectory}).

\section{Transient dynamics across linear model families}
\label{sec:transient}

We compare progressively richer linear descriptions of per-position updates
along the transient and identify the streams they predict least well.
Sec.~\ref{sec:intervention} evaluates interventions at those sites.

\subsection{Excluding time-invariant, second-order, and coarse-clock linear descriptions}
A single time-invariant operator $\dz_{t+1}\!\approx\!A\dz_t$ (streaming least-squares
$A{=}C_1C_0^{+}$, uncentred one-step $R^2$) fits poorly at per-position resolution,
explaining only $R^2{=}0.43/0.18/0.12$ of the update variance for $\zL/\zH/\zA$;
position-averaging inflates these to $0.90/0.49/0.38$, a pooling artifact
(App.~\ref{app:synthetic}). The residual is as high-rank as the signal, so no
low-dimensional dropped modes recover it, and a second-order term ($\dz_{t+1}$ from
$[\dz_t,\dz_{t-1}]$) captures only $5$--$10\%$ of it: linear memory does not account
for it either. These tested linear families leave substantial residual variation.

The recurrence has an explicit clock, so the operator could be piecewise-linear,
switching by phase; but conditioning on a \emph{coarse} phase recovers almost
nothing and yields operators stable across puzzles yet drifting across trajectory
time, superficially the fingerprint of state-dependent nonlinearity. This is a
trap: trajectory time \emph{is} the pooled-away clock, so a fully scheduled but
purely \emph{linear} $A(s,h,\ell)$ reproduces the signature with no state
dependence (App.~\ref{app:synthetic}).

\subsection{The full-clock test leaves the slow loop substantially less linearisable}
\label{sec:fullclock}
We close the loophole by bucketing every update by its complete clock
$(s,h,\ell)$ (segment, H-cycle, L-step), yielding $62$ $\zL$ and $30$ $\zH$
per-position operators ($64$/$32$ snapshots give $63$/$31$ updates, and pairing
consecutive updates leaves $62$/$30$ buckets, a structural count that precedes
any energy or reliability filter; App.~\ref{app:buckets}), each fit on $\geq$41{,}984 vectors ($\approx$81:1). Because
the noise-dominated near-fixed-point tail swamps an unweighted median, we summarise by
the energy-weighted mean over the top-tercile ticks $\mathcal{T}$ (per-tick energy
$E_\varphi{=}\sum_{t\in\varphi}\lVert\dz_t\rVert^2$),
\begin{equation}
R^2_{\mathrm{hi}} = \frac{\sum_{\varphi\in\mathcal{T}} E_\varphi\,R^2_\varphi}{\sum_{\varphi\in\mathcal{T}} E_\varphi},
\label{eq:hienergy}
\end{equation}
and de-bias for conditioning by a split-half check: each puzzle-half's operator
scored on the other half,
\begin{equation}
R^2_{\mathrm{ho}}{=}\tfrac12\!\big[R^2(A_{\mathrm e};\mathrm{o}){+}R^2(A_{\mathrm o};\mathrm{e})\big],\ \
R^2_{\mathrm{corr}}{=}R^2_{\mathrm{ho}}/R^2_{\mathrm{in}}.
\label{eq:heldout}
\end{equation}

\begin{figure*}[t]
\centering
\includegraphics[width=\linewidth]{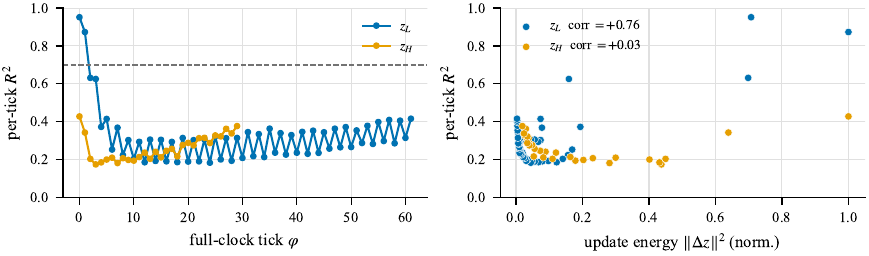}
\caption{Full-clock per-tick linear fits, solved stratum, \emph{raw} per-tick
$R^2$. \textbf{Left:} $R^2$ along the clock $\varphi$ ($62$ $\zL$ ticks against
$30$ $\zH$). \textbf{Right:} the same $R^2$ against that tick's update energy
$\|\dz\|^2$. The fast $\zL$ recurrence's linearity \emph{rises} with energy
(Pearson $+0.76$, peak $0.95$), so the energy-weighted summary is carried by the
ticks that do the work; the slow $\zH$ shows no such structure ($+0.03$) and
stays weakly fit even where it is most energetic. The split-half-corrected
energy-weighted figures ($0.72$ against $0.37$) and the estimator behind them are
in App.~\ref{app:buckets}.}
\label{fig:fullclock}
\end{figure*}

\newcommand{\fullclocktable}{%
\begin{table}[t]
\centering
\caption{Full-clock per-tick linear fits (per-position). hi-E is the
energy-weighted $R^2$ over the top-tercile ticks (Eq.~\ref{eq:hienergy}); held-out
and $R^2_{\mathrm{corr}}$ are the split-half reliability figures
(Eq.~\ref{eq:heldout}; ${\approx}1$ = reliably estimated). The estimator behind
every cell, including the order its two selection steps run in, is
Algorithm~\ref{alg:fullclock}. The time-invariant-operator
$R^2$ is the first rung of Fig.~\ref{fig:overview}(b) and is quoted in
Sec.~\ref{sec:transient}; the raw ceiling enters only through
$R^2_{\mathrm{corr}}$, so neither is repeated here.}
\label{tab:fullclock}
\begin{tabular}{@{}lcccc@{}}
\toprule
stratum & hi-E & \textbf{held-out} & $R^2_{\mathrm{corr}}$ & corr$(E,R^2)$ \\
\midrule
solved $\zL$   & 0.73 & \textbf{0.72} & 0.96 & $+0.76$ \\
solved $\zH$   & 0.39 & \textbf{0.37} & 0.90 & $+0.03$ \\
unsolved $\zL$ & 0.80 & \textbf{0.80} & 0.99 & $+0.80$ \\
unsolved $\zH$ & 0.42 & \textbf{0.39} & 0.91 & $+0.88$ \\
\bottomrule
\end{tabular}
\end{table}}
\ificlr\else\fullclocktable\fi

\textbf{Result} (Fig.~\ref{fig:fullclock}). The fast recurrence's linearity
concentrates where the energy is: its top-3 ticks carry $51\%$ of all update
energy. The concentration is not a monotone trend (Pearson $+0.76$, rank
$-0.29$), which we report because the same rank test demotes a claim below.
Split-half correction leaves the gap intact (held-out $+0.35$ vs.\ raw $+0.33$;
App.~Fig.~\ref{fig:reliability}), so in-sample optimism does not produce it.
Held-out $R^2$ still tracks each bucket's signal-to-noise ratio and does not
separate unpredictable variation from nonlinearity. The one
apparent asymmetry (for $\zH$, energy--linearity alignment $+0.03$ solved vs.\
$+0.88$ unsolved) is a fixed-point-tail effect plus a leverage artifact
(App.~\ref{app:buckets}). The robust difference is dynamical:
unsolved sustains high-energy poorly-captured updates late (segment ${>}8$: median energy
${\approx}4.5\times$ solved's, $R^2{\approx}0.21$), consistent with continued
search, while solved's updates have decayed.

\textbf{Conditioning on the input drive.}
Adding the injection $z_H+x$ to the per-tick regression does not improve
held-out prediction at the two highest-energy $z_L$ ticks
($\Delta R^2=-0.003$; App.~\ref{app:injection}).
This supports the predictive contribution of the previous update, but
redundant predictors do not establish causal independence from the drive.

\textbf{Scope of the timescale comparison.}
The tested predictors leave more residual variation in HRM's slow $z_H$
updates than in its fast $z_L$ updates. Kernel and MLP refits recover
approximately $0.03$ additional held-out $R^2$
(App.~\ref{app:nonlinear}). This motivates the site comparison in
Sec.~\ref{sec:intervention}, but does not identify a causal role for the
residual. Across-puzzle operator drift is dominated by conditioning and is
not used as a state-dependence diagnostic (App.~\ref{app:drift}).

The default slow-loop estimates use only $1$--$3$ reliable high-energy
ticks. Relaxing the energy and reliability thresholds retains $10$--$14$
ticks while preserving the HRM ordering (App.~\ref{app:sensitivity}).
Composing fast updates onto the slow clock removes $37\%$ of the solved
gap, leaving $0.22$ (App.~\ref{app:stepsize}).
The ordering recurs across HRM seeds and attenuates on the published Maze
checkpoint, though not on two others (App.~\ref{app:maze-clean}), and reverses
on unsolved TRM states. It is a property of the measured settings and
estimator, not a universal division of computation.

\section{Relation to constraint propagation}
\label{sec:compute}

The later iterations carry the harder part of the computation. Running a
naked/hidden-singles propagator to fixation on each solved puzzle, the segment at
which the model's lens commits a cell---the first segment whose H-lens argmax for
that cell is correct and stays correct to the last---tracks its propagation depth
(within-puzzle Spearman $\rho{=}0.32$). The model does more than propagate: only
$20.9\%$ of solved puzzles are fully singles-determined, and the cells the
propagator leaves unforced resolve substantially later than the forced ones
($6.3$ vs.\ $2.1$). Search-like work thus concentrates in the later iterations,
where the fits are weakest (Sec.~\ref{sec:transient}; counts and tests in
App.~Figs.~\ref{fig:propagation},~\ref{fig:casestudy}).

\section{Reselection and compute-matched alternatives}
\label{sec:intervention}

Perturbation and reselection can generate and select alternative predictions
\citep{efstathiou2026,ptrm2026,corbett2026}.
This experiment differs from the persistence test: it injects after five of
$16$ segments (zero-based index $4$) and scores correctness, whereas Sec.~\ref{sec:fail} continues
terminal states and scores agreement with their original answers.
High terminal persistence does not preclude reselection gains, as
GRAM-guided TRM demonstrates (App.~\ref{app:gram}).

\begin{figure*}[t]
\centering
\includegraphics[width=\figw]{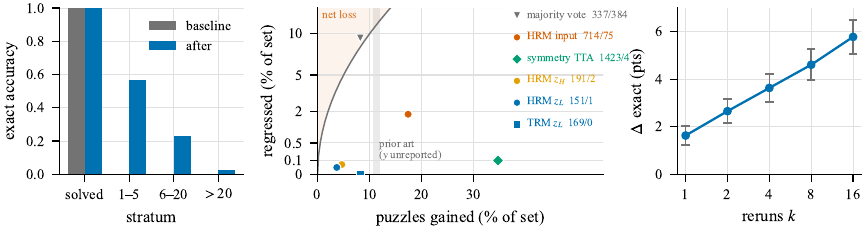}
\caption{Oracle-free noise reselection, $k{=}8$, reselected by the model's own
halting logit. \textbf{Left:} exact accuracy by stratum, balanced $1024$-puzzle
sample; the gain decays with severity ($+56$, $+23$, $+3$ points). Uniform
$N{=}4096$: $+4.76$, paired $95\%$ CI $[+4.10,+5.42]$. \textbf{Middle: the
accounting, which is what we claim.} Gained vs.\ \emph{regressed} puzzles
(fraction of set, $\sqrt{\cdot}$ on $y$; solid $=$ break-even); sites are at
unmatched $\varepsilon$, and App.~\ref{app:ablation} matches their radii.
Prior work reports no regression count, so the band gives only the constraint implied by its net gain.
\textbf{Right:} monotone in budget ($+1.6\!\to\!+5.8$, $k{=}1\!\to\!16$); the
$k{=}8$ point ($+4.6$) is an independent draw of the same puzzles.}
\label{fig:intervention}
\end{figure*}

\textbf{Four procedures, one selector} (Fig.~\ref{fig:intervention}; uniform
$N{=}4096$, all oracle-free, reselected by the model's own halting logit). Injecting noise into $\zH$
mid-run gives $+4.8$ points ($197$ gained, $2$ lost). The \emph{same} $9\times$
budget spent on depth alone---running the recurrence to $144$ segments with no
perturbation anywhere, $9\times$ the trained horizon and so an extrapolation
(App.~\ref{app:depth})---gives $+5.6$ ($245/14$),
and reading out the max-q-halt
segment of that single pass gives $+6.0$ ($248/2$; App.~\ref{app:depth}).
These procedures match total recurrent computation but differ in available
parallelism and potential latency; App.~\ref{app:depth} specifies the
extended-horizon implementation and its extrapolation limits.
A task-specific symmetry augmentation gives $+34.6$ ($1423/4$) and largely
\emph{subsumes} the injection: pooling all $17$ candidates adds only $+0.51$ over
symmetry alone, with selected accuracy consistent with the candidate oracle ceiling
(nine-orientation oracle union $76.5\%$;
App.~\ref{app:tta}). We therefore do not claim noise injection as
the best use of a test-time budget.
Depth corrects $245$ baseline errors and introduces $14$ regressions,
demonstrating that some terminal errors remain recoverable by continuation.
Compute matching does not imply latency matching: the $144$-segment trajectory
is sequential, whereas the $k{=}8$ perturbed continuations can run in parallel
after a shared prefix, subject to hardware and memory constraints. Parallel
execution trades additional resources for potentially lower latency; we do
not measure hardware-specific wall-clock performance.

\textbf{Candidate generation versus ranking.}
At $z_H$, q-halt recovers $99.0\%$ of the improvement available in the
oracle union (the set of puzzles solved by at least one candidate).
Under site-relative scaling, input injection creates a larger oracle union
($0.484$ versus $0.462$), but selection yields a net loss and recovers
$-16\%$ of its headroom (App.~\ref{app:ablation}).
The input-site failure is therefore a ranking limitation; distribution
shift in the scored states is a possible explanation.
Within-puzzle q-halt spread differs between recoverable and unrecoverable
errors (App.~\ref{app:disagreement}), but does not alone explain selection.
Per-cell majority voting over noisy reruns loses $0.51$ points.

\textbf{Where to inject, and where that stops.} Two lost against one out of
$4096$ carries no statistical content, so we scored both recurrent sites on the
\emph{full} $422{,}530$-puzzle frame. The rates separate, and not in the
direction the timescale split would suggest: $\zH$ regresses $0.0090\%$ (Wilson
$[0.0066,0.0123]$) against $\zL$'s $0.0036\%$ $([0.0022,0.0059])$, $2.5\times$ as
often. The timescale split therefore picks the \emph{higher-gain} site
($+4.36$ vs.\ $+2.91$), not the safer one; among recurrent sites there is a trade
(App.~\ref{app:ablation}). Both have much lower observed regression rates than the input
embedding, net negative at matched magnitude. That localization does not survive module sharing: in TRM, injecting
into either loop gives the same lift ($+8.25$ vs.\ $+8.40$; App.~\ref{app:trm}).
Protocol, $k$-sweep, timing and per-site scales: App.~\ref{app:ablation}.

\section{Discussion and Limitations}
\label{sec:discussion}

\paragraph{Architectural vs.\ computational hierarchy.}
Collapsing HRM's two levels into one shared recurrence costs no accuracy
\citep{trm2025}, so the \emph{architectural} split is droppable; the
\emph{computational} one is what isolates the least-linearisable work, and in HRM
the higher-gain injection site with it. Under module sharing that split does not
vanish, it \emph{inverts}: the fast $\zL$ loop becomes the less-linearisable
stream ($R^2{=}0.45$ against $0.62$, unsolved stratum; TRM's solved stratum does
not invert, $0.48/0.48$), and the intervention still transfers---$+8.25$ ($169/0$)
into $\zL$, an indistinguishable $+8.40$ ($172/0$) into $\zH$. The method still
isolates the loop; only in HRM does that localization also say where to intervene
(Apps.~\ref{app:trm},~\ref{app:ablation}).

\paragraph{Implications and limitations.}
Continued depth and diverse reruns can correct some terminal errors, while
the halting head recovers much of the available candidate headroom.
A nonlinear trajectory readout also improves correctness discrimination
(App.~\ref{app:nonlinear-readout}).

The main geometric results concern one $27.3$M-parameter HRM checkpoint,
so they are best read as a case study of this model, not as properties of
recurrent reasoners in general.
A second seed preserves the reported directions. Maze remains a
reduced-scale replication: its unsolved fast--slow gap changes from $0.37$
to $0.04$, and its endpoint spectral gap is inconclusive on $107$ solved
mazes. Full-clock refits also find a positive gap without a warm restart;
differences between training runs prevent isolating the cause of its
changed magnitude (App.~\ref{app:maze-clean}).
TRM's spectral ordering disappears under the terminal-update gate, and
its timescale inversion is confined to unsolved states.
Low held-out $R^2$ establishes limited prediction by the tested families,
not a particular computational mechanism or irreducible nonlinearity.
The coverage figure now has stored metrics for every cell; the separate
third-party GRAM endpoint pilot still requires external materials
(App.~\ref{app:gram}).

\section{Conclusion}
Wrong terminal answers in the primary HRM checkpoint are less persistent than
correct answers under continued recurrence. An unperturbed control shows that
the tested perturbations contribute little to this difference.
The solved--wrong ordering extends across the evaluated settings, whereas
absolute thresholds, endpoint spectra and timescale geometry do not.
Noise, extended depth and task symmetries provide distinct sources of
alternative predictions; paired gains, regressions and oracle headroom
clarify their benefits and limitations. These results characterise finite-horizon
failure dynamics without excluding asymptotic attractors.

\newcommand{\refsthenappendix}{%
  \ificlr\bibliographystyle{iclr2027_conference}\else\bibliographystyle{icml2026}\fi
  \bibliography{references}%
  \appendix
  \raggedbottom}
%
\ificlr
\section*{AI Use Statement}
LLM assistants were used to implement synthetic generators, analysis code,
tests and plotting scripts; develop diagnostic criteria and controls;
search and critique related work; and draft and revise the manuscript.
Puzzle instances came from released dataset builders, not LLM generation.
The paper contains no theorems or proofs.

Primary HRM results were computed by analysis code and checked against its
stored outputs. Incomplete artifact coverage for additional analyses is
identified in the corresponding appendices. Bibliographic records dated
2025 or later were checked against source records. Detailed descriptions of
\citet{ptrm2026,corbett2026,renliu2026} were checked against their full texts;
\citet{efstathiou2026} is quoted from its published abstract, its full text
being unavailable (App.~\ref{app:disagreement}). These checks do not imply independent human
re-derivation of every statement. The authors directed the research and take
responsibility for the complete paper, including any errors. No LLM is an author.
\fi
\ificlr\section*{Ethics Statement}\else\section*{Impact Statement}\fi
This study analyses small recurrent models on algorithmically generated
constraint-satisfaction puzzles. It involves no human subjects, personal data
or sensitive data. Its results do not establish transfer to deployed systems.

A halting head that ranks candidates well can still misclassify individual
outputs, so its confidence alone is insufficient to establish correctness.
Repeated inference also incurs additional compute: our noise-injection
procedure uses $9\times$ evaluation compute for a $4.8$-point gain, while
the compute-matched Sudoku symmetry baseline gains $34.6$ points.
Applications should compare such alternatives before adopting the procedure.
\ificlr\else\refsthenappendix\fi

\section*{Reproducibility Statement}
The primary checkpoint and training configuration are specified in
Sec.~\ref{sec:setup}. Evaluation uses fixed manifests for the stratified
$512{+}512$ sample, the disjoint uniform $4096$-puzzle sample, and the
$32{+}32$ full-trajectory cross-checks. Recorded activations are stored in
\texttt{fp32}; the streaming second moments and linear-operator fits accumulate
in \texttt{fp64}; streaming and direct fits agree to
$\max|\Delta R^2|=1.14\times10^{-14}$. The instrumented inference loop is
checked against the model forward pass to \texttt{bfloat16} tolerance, and the
test suite covers the analysis and inference checks. Algorithm~\ref{alg:fullclock} specifies the
linearity estimator; App.~\ref{app:relabel} documents replay labels, inclusion
probabilities and weighted bootstraps; App.~\ref{app:ablation} gives the frozen
intervention protocol; and App.~\ref{app:prereg} distinguishes pre-specified
analyses from exploratory ones. The model-independent analysis subpackage is
tested to import without a deep-learning framework. Stored-artifact gaps for
nonlinear readouts, replications and pilots are disclosed in their appendices.
Full residual-stream traces are not persisted; reproducing analyses that lack
stored metrics requires rerunning the checkpoint.
\ificlr\refsthenappendix\fi

\section{Synthetic regression tests for the two artifacts}
\label{app:synthetic}
Both methodological artifacts documented in the main text are pinned by
synthetic regression tests in our suite; here we describe the
constructions and their expected signatures.

\begin{figure*}[t]
\centering
\includegraphics[width=\linewidth]{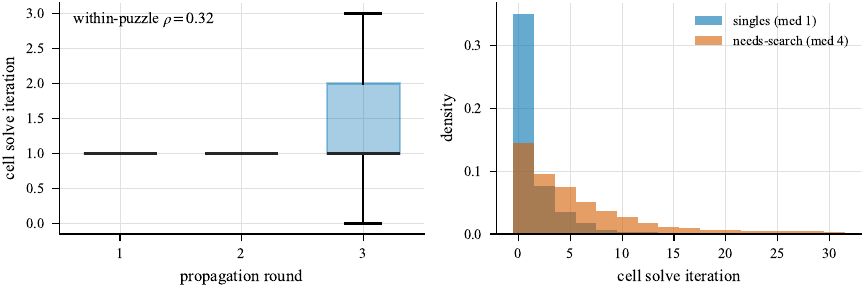}
\caption{Constraint-propagation grounding (solved stratum). Cell solve-time tracks naked/hidden-singles propagation depth (within-puzzle Spearman $\rho{=}0.32$); search-requiring cells resolve later than singles-determined ones (mean iteration $6.3$ vs.\ $2.1$, Mann--Whitney $p$ below reported numerical precision). The obvious objection is that
search-requiring cells are simply \emph{harder}; matching cells on their
surviving-candidate count within each puzzle leaves the gap at $+2.7$ segments
($19/19$ puzzles positive, two-sided exact sign test
$p{=}3.8\!\times\!10^{-6}$; the unmatched comparison App.~\ref{app:synthetic}
quotes is a different, one-sided test at $18/19$), and a cell-level ANCOVA over
$1{,}740$ cells gives the needs-search indicator $+2.26$ segments against $+0.12$
per additional candidate. Its standard errors, $0.15$ and $0.06$, are ordinary
least squares over cells and treat the cells of one puzzle as independent, so
they are narrower than a puzzle-clustered interval. Difficulty is real and small;
it is not what the fingerprint is tracking. Only $20.9\%$ of solved puzzles are fully
singles-determined; the model resolves $19{,}352$ cells the propagator cannot.}
\label{fig:propagation}
\end{figure*}

\paragraph{Solve-timing control.}
Solve-timing comparisons use solved puzzles to avoid right-censoring cells
that never settle. Needs-search cells resolve later in $18/19$ puzzles
(one-sided sign test $p=3.8\times10^{-5}$). Matching on candidate count gives
$19/19$ positive differences (two-sided $p=3.8\times10^{-6}$;
Fig.~\ref{fig:propagation}); these are distinct tests.

\textbf{(a) Mean-field inflation.} We generate a multi-position linear system
whose per-position updates share an operator $A$ but carry an independent
mean-zero per-position deviation $B_s$, then fit $\dz_{t+1}\!\approx\!A\,\dz_t$
two ways. A single shared operator is the right estimand, and the honest
per-position fit recovers it only partly; averaging positions before fitting
drives the fit to saturation and raises the complex mass with it. Both
inflations reproduce over seeds, in the direction of the empirical jump ($\zL$
$R^2$ $0.43\!\to\!0.90$, complex mass ${\sim}0.63\!\to\!{\sim}0.90$; the honest
per-position complex mass is the one that matches the $65\%$ complex fraction of
the endpoint Jacobians' retained Ritz values, App.~\ref{app:jacobian}); the
synthetic's \emph{magnitudes} follow its own dimension and are not claimed to
match. Two mechanisms travel under one name here. Pooling suppresses position-specific
structure by ${\sim}S^{-1/2}$, and it \emph{also} divides the
sample-to-dimension ratio by $S$---here from ${\sim}70{:}1$ to under $1{:}1$---so
the pooled fit is underdetermined, its in-sample $R^2$ saturates, and its
pseudo-inverse manufactures spurious oscillatory modes. The $B_s{=}0$ control
lifts the honest fit to the same ceiling, so it cannot assign the $R^2$
inflation to averaging rather than to the collapsed pair count. The synthetic
pins the direction of both inflations; it does not separate their causes.
Per-position reduction is the correct default for both reasons.

\textbf{(b) Under-specified clock.} We construct a three-level-clock scheduled
linear dynamical system (a fully \emph{state-independent} program $A(\text{seg},H,L)$) and
condition it on only a coarse phase while treating the pooled-away clock levels
as ``time.'' This reproduces the exact false state-dependence signature the coarse-phase
test of Sec.~\ref{sec:transient} returns: low coarse-phase $R^2$ together with
high drift-across-time and low drift-across-puzzles. Conditioning on the \emph{full} clock then resolves it
($R^2\!\approx\!1$, across-puzzle drift $\approx\!0$), confirming that the
signature is a clock artifact rather than evidence of nonlinearity.

\section{Controls for operator drift and expanding modes}
\label{app:drift}
The even/odd across-puzzle operator drift cannot serve as a state-dependence
probe. Each half is a single ${\sim}21$k-pair fit, and a synthetic
well-conditioned null places the estimation floor at $R^2$-scale drift
${\sim}0.04$--$0.17$, rising to ${\sim}0.6$ in the ill-conditioned
near-fixed-point regime. The observed drift anti-correlates with per-tick update
energy ($\mathrm{corr}(E,\text{drift})=-0.86$ and $-0.99$ for $\zL$/$\zH$),
exactly the conditioning-driven pattern: the noisiest, lowest-energy ticks
inflate drift. We therefore retire drift as a state-dependence signal and rely
instead on the split-half held-out $R^2$ of Sec.~\ref{sec:fullclock}
(Eq.~\ref{eq:heldout}), which measures fit quality after de-biasing for SNR and
conditioning.

\begin{figure*}[t]
\centering
\includegraphics[width=\linewidth]{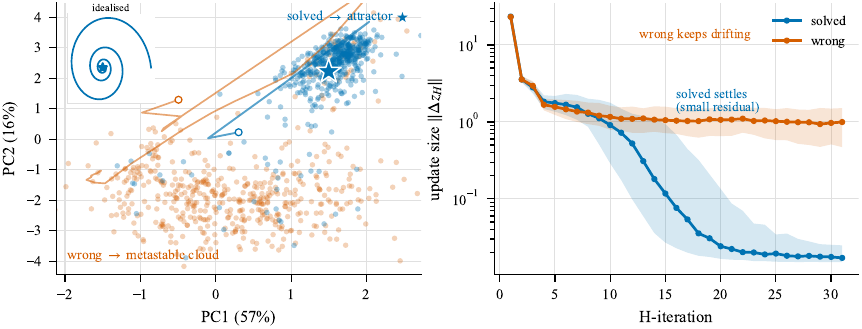}
\caption{Solved and wrong $z_H$ trajectories.
\textbf{Left:} the first two PCA components of position-averaged states,
shown only for orientation. Neither clustering nor apparent rotation in this
projection establishes a fixed point or its stability
(App.~\ref{app:synthetic}). All runs share the learned initial state.
\textbf{Right:} per-position update norms (median and IQR) decay toward zero
for solved trajectories but remain larger for wrong trajectories.
Endpoint spectra and answer persistence are measured separately in
Secs.~\ref{sec:endpoint} and~\ref{sec:fail}.}
\label{fig:trajectory}
\end{figure*}

\paragraph{The expanding mean-field mode.}
\phantomsection
\label{app:expanding}
A late-window position-mean $\zA$ operator yields a spectrally expanding,
complex-oscillatory mode (dominant $\lambda=-0.71\pm0.89j$, $\rho=1.14$,
$R^2=0.94$). It is however \emph{window-fragile}: $\rho$ falls monotonically as
the fit window widens ($w{=}3\!:\!1.27$, $R^2\,0.99$; $w{=}5\!:\!1.14$, $0.94$;
$w{=}7\!:\!0.89$, $0.77$; $w{=}10\!:\!0.66$, $0.50$; $w{=}15\!:\!0.56$, $0.37$),
crossing unity only in the last-third window. The mode is genuinely
readout-enriched: its row-space fraction on the $11$-dimensional readout is
$0.36$ against a chance level of $0.021$, a $17\times$ enrichment (the $q$-head
read fraction is $0.10$). But the ``confidence-sharpening'' attribution does not
survive a random-direction null: the mode's margin correlation ($r=0.69$) sits
below the null $95$th percentile ($0.70$), and the net-drift direction correlates
more strongly ($0.90$). We therefore report the enrichment as solid and the
sharpening interpretation as \emph{not yet established}.

\paragraph{Matched perturbation controls.}
We perturb along the fitted mode at zero-based injection index $10$, after
eleven of the $16$ segments, with Frobenius norm
equal to the median final-segment update. Controls use the same code path:
$64$ isotropic directions, $24$ subdominant eigenvectors of the same operator,
and $64$ random directions in the top-$16$ principal subspace of state deltas.
The latter two control for alignment with the anisotropic state distribution.

Final-state displacement is $0.055$ of the state norm for the mode, versus
$0.054$ for subspace controls, $0.058$ for eigenvector controls, and $0.036$
for isotropic controls. The mode's answer-change rate exceeds all isotropic
controls on all four tested axes and both signs, but lies at only the
$25$th--$63$rd percentile of alignment-matched controls.
Its margin effect lies at quantile rank $0.016$--$0.812$ within the matched
families and has inconsistent signs across axes. Thus the tested intervention
does not support a distinctive confidence-sharpening role, despite readout
enrichment. An earlier isotropic-only comparison gave a misleading positive
result; regression tests now enforce all three control families.

\section{Excluded buckets and conditioning details}
\label{app:buckets}
\ificlr\fullclocktable\fi

\begin{algorithm}[t]
\small
\caption{Full-clock linearity estimate for one (stratum, timescale).
$R^2(A;X)$ scores operator $A$ on half $X$; the per-tick loop computes
Eq.~\ref{eq:heldout} and the return is Eq.~\ref{eq:hienergy}. Steps are named
rather than numbered because the numbers move when the listing is edited: the
listing below is the procedure we ran, and no other.}
\label{alg:fullclock}
\begin{algorithmic}[1]
\STATE {\bfseries Input:} updates $\dz_t$, each tagged with its full clock tick
       $\varphi{=}(s,h,\ell)$ and its puzzle id
\STATE {\bfseries Params:} $c{=}\tfrac13$, $\tau{=}0.3$ (App.~\ref{app:sensitivity})
\FOR{each clock tick $\varphi$}
  \STATE split puzzles by parity into halves $\mathrm e,\mathrm o$
  \STATE accumulate $(C_0,C_1,S_{yy})$ per half; $A_X\!\leftarrow\!C_1^X(C_0^X)^{+}$
  \STATE $R^2_{\mathrm{ho}}\!\leftarrow\!
         \tfrac12[R^2(A_{\mathrm e};\mathrm o)\!+\!R^2(A_{\mathrm o};\mathrm e)]$
  \STATE $R^2_{\mathrm{ceil}}\!\leftarrow\!
         \tfrac12[R^2(A_{\mathrm e};\mathrm e)\!+\!R^2(A_{\mathrm o};\mathrm o)]$
  \STATE $E(\varphi)\!\leftarrow\!\sum_{t\in\varphi}\lVert\dz_t\rVert^2$
\ENDFOR
\STATE $\Phi_{\mathrm{fin}}\!\leftarrow\!
       \{\varphi: R^2_{\mathrm{ceil}}(\varphi)\ \text{finite}\}$
\STATE \textbf{rank:} $\Phi_{\mathrm{top}}\!\leftarrow\!$ the
       $\max(1,\mathrm{round}(c|\Phi_{\mathrm{fin}}|))$ ticks largest in $E$
\STATE \textbf{filter:} $\Phi_{\mathrm{use}}\!\leftarrow\!\{\varphi\!\in\!
       \Phi_{\mathrm{top}}: R^2_{\mathrm{ceil}}(\varphi)\!\ge\!\tau\}$
\STATE report the excluded $\Phi_{\mathrm{top}}\!\setminus\!\Phi_{\mathrm{use}}$
\STATE {\bfseries return} $R^2_{\mathrm{hi}}\!=\!
       \sum_{\Phi_{\mathrm{use}}}\!E R^2_{\mathrm{ho}}
       \big/\!\sum_{\Phi_{\mathrm{use}}}\!E$
\end{algorithmic}
\end{algorithm}

\textbf{Algorithm~\ref{alg:fullclock}'s \emph{rank} and \emph{filter} steps do not commute}, which is the reason this procedure is
written out rather than described. Ranking first and filtering second reports the
energetically dominant computation, minus whichever of those ticks cannot be
scored reliably, and says which. Filtering first would admit a low-energy but
well-conditioned tick into the top-$c$ in its place, so the number would describe
a different computation while carrying the same name. Our implementation fixes the
order in one place and a unit test pins it; the two orders are not otherwise
distinguishable from a reported $R^2_{\mathrm{hi}}$.

Full-clock buckets whose de-biased ceiling falls below $0.3$ are excluded from
the headline hi-energy fits. These are the low-energy, near-fixed-point ticks:
once the trajectory has contracted, per-tick updates are dominated by noise and
the fit is ill-conditioned, so any $R^2$ there reflects estimation error rather
than dynamics. The retained buckets are well-specified: each full-clock operator on a full
stratum is estimated from $41{,}984$ update vectors ($512$ puzzles $\times$ $82$
positions), $82$ observations per predictor. The held-out operators of
Eq.~\ref{eq:heldout} are each fitted on one parity half, $20{,}992$ vectors, so
their ratio is $41\!:\!1$. As a correctness check, the streaming
second-moment estimator agrees with a direct fit on the held full trajectories to
$\max|\Delta R^2| = 1.1\times10^{-14}$, confirming that streaming introduces no
numerical bias.

\begin{figure*}[t]
\centering
\includegraphics[width=\linewidth]{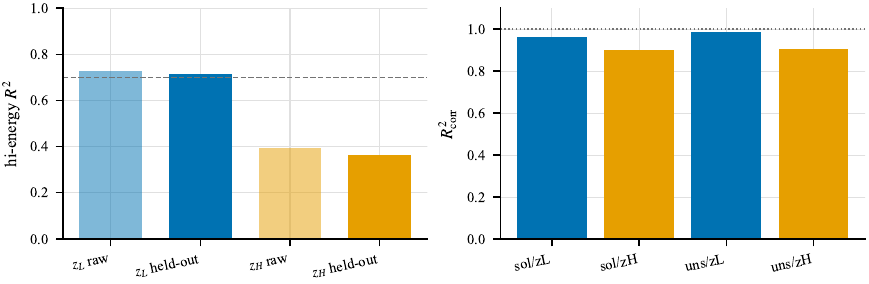}
\caption{Split-half reliability (Sec.~\ref{sec:transient}).
\textbf{Left:} raw vs.\ held-out hi-energy $R^2$ for $\zL$/$\zH$: the fast/slow
gap survives the correction (held-out $0.72$ vs.\ $0.37$, a $+0.35$ gap).
\textbf{Right:} $R^2_{\mathrm{corr}}$ per timescale and stratum (${\approx}1$ =
reliably estimated).}
\label{fig:reliability}
\end{figure*}

\begin{figure*}[t]
\centering
\includegraphics[width=\linewidth]{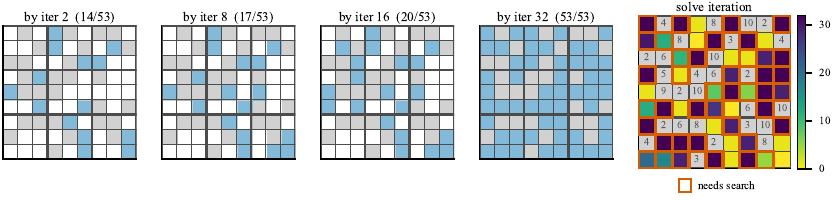}
\caption{\textbf{One solved Sudoku (illustrative).} \textbf{First four panels:}
cumulative solve-time wavefront: a cell is filled once the H-lens fixates it
correctly by iteration $t$ (count shown); a core resolves early, the rest commit
late. \textbf{Right:} the same board coloured by solve iteration (light early, dark
late), with needs-search cells (no naked/hidden single; red outline) tending to
resolve later than singles-determined ones, the single-puzzle view behind the
aggregate $\rho{=}0.32$ of Fig.~\ref{fig:propagation}.}
\label{fig:casestudy}
\end{figure*}

\section{Input-conditioned and nonlinear refits}
\label{app:injection}
The fast-loop linearity claim (Sec.~\ref{sec:transient}) faces one alternative:
the L-module is fed the injection $\zH{+}x$ each step, so a per-tick operator could
be absorbing that exogenous linear input rather than performing genuine recurrent
dynamics. We test this directly. Each full-clock tick is refit jointly on the
previous update and the injection,
\[
  \dz_{t+1} \approx A_\phi\,\dz_t + B_\phi\,(\zH{+}x) + c ,
\]
and we score three nested sub-models held-out across a puzzle-parity split (fit on
even-indexed puzzles, score on odd, and vice versa): state-only
$[\dz_t,\,1]$, injection-only $[\zH{+}x,\,1]$, and full
$[\dz_t,\,\zH{+}x,\,1]$. All three carry the intercept $c$, so the state-only
sub-model is affine and is not the through-origin fit of
Sec.~\ref{sec:fullclock}. The operator's \emph{unique} held-out
gain is $\Delta_{\text{op}} = R^2_{\text{full}} - R^2_{\text{inj}}$, the
variance the recurrent operator explains that the injection cannot; symmetrically
$\Delta_{\text{inj}} = R^2_{\text{full}} - R^2_{\text{state}}$. Held-out scoring is
essential: adding the injection block can only inflate the in-sample fit, so the
sign of $\Delta_{\text{inj}}$ is meaningful only out of sample.

Table~\ref{tab:injection} reports the three most energetic $z_L$ ticks on
the stratified $1024$-puzzle sample. At the top two, adding the injection
changes held-out $R^2$ by $-0.003$: it adds no predictive information given
the previous update. At the third, lower-energy tick, it improves prediction.
This pattern occurs in both strata; no comparable operator dominance appears
for $z_H$. Predictor redundancy does not establish that the drive is
causally irrelevant.

\begin{table}[h]
\centering
\small
\caption{Injection-conditioned refit at the three most energetic $\zL$ ticks
(held-out $R^2$; ticks ordered by update energy). $\Delta_{\text{op}}$ is the
operator's unique gain over the injection, $\Delta_{\text{inj}}$ the injection's over
the operator. At the two dominant ticks the injection adds nothing
($\Delta_{\text{inj}}\!\approx\!0$).}
\label{tab:injection}
\fittab{%
\begin{tabular}{llccccc}
\toprule
stratum & tick & state & inj & full & $\Delta_{\text{op}}$ & $\Delta_{\text{inj}}$ \\
\midrule
solved   & 1 & $0.872$ & $0.730$ & $0.868$ & $+0.138$ & $-0.003$ \\
solved   & 0 & $0.951$ & $0.898$ & $0.948$ & $+0.050$ & $-0.003$ \\
solved   & 2 & $0.625$ & $0.807$ & $0.818$ & $+0.010$ & $+0.193$ \\
\midrule
unsolved & 1 & $0.873$ & $0.729$ & $0.869$ & $+0.140$ & $-0.003$ \\
unsolved & 0 & $0.951$ & $0.898$ & $0.949$ & $+0.050$ & $-0.003$ \\
unsolved & 2 & $0.664$ & $0.820$ & $0.829$ & $+0.010$ & $+0.165$ \\
\bottomrule
\end{tabular}}
\end{table}

\paragraph{Nonlinear refits recover little beyond the linear fit.}
\label{app:nonlinear}
To test whether $\zH$'s low per-tick linear $R^2$ reflects a \emph{recoverable}
nonlinearity or genuine complexity/noise, we refit the same per-tick, per-position
problem ($\dz_{t+1}=g(\dz_t)$) on the top-tercile energy ticks with two nonlinear
classes: RBF kernel ridge (Nystrom, ${\sim}2048$ landmarks) and a
$512{\to}1024{\to}512$ GELU MLP (early-stopped on a validation split), under the
same split-half (puzzle-parity) protocol as Table~\ref{tab:fullclock} but aggregated over
\emph{all} top-tercile ticks (no reliability-ceiling exclusion) and fit on a
capped per-tick pair reservoir, so all three classes see identical pairs; this
lowers the absolute linear baseline relative to Table~\ref{tab:fullclock} (read
within-row; see the Table~\ref{tab:nonlinear} caption). Both lift
$\zH$ by only ${\approx}0.03$ held-out $R^2$ over linear, comparable to their gain on
$\zL$, and neither approaches a good fit. The method is not at fault: on a synthetic
\emph{state-dependent} generator it recovers $+0.27$ over linear, and on the
scheduled-\emph{linear} control (App.~\ref{app:synthetic}) it correctly recovers
${\approx}0$. The residual therefore remains poorly predicted by these particular kernel
and MLP fits. Reproducible held-out performance reduces concern about
estimation error in the fitted operators, but does not distinguish
unobserved inputs, stochastic variation and more complex nonlinear structure.

\begin{table}[h]
\centering \small
\caption{Held-out $R^2$ of per-tick refits on the top-tercile energy ticks (same
split-half protocol as Table~\ref{tab:fullclock}). Nonlinear classes barely exceed linear on the real
data; the synthetic controls confirm the method's sensitivity. \textbf{Read
within-row.} The \emph{linear} column is lower than Table~\ref{tab:fullclock}'s
held-out hi-E ($\zL$ $0.55$ vs.\ $0.72$; $\zH$ $0.12$ vs.\ $0.37$) for two reasons,
both applied identically to all three classes: (i) this tercile mean retains
\emph{all} top-tercile ticks, whereas Table~\ref{tab:fullclock} additionally drops
ticks whose split-half reliability ceiling is $<\!0.3$ (near-fixed-point,
rank-collapsed), the dominant factor ($\zL\,0.72{\to}0.58$, $\zH\,0.37{\to}0.20$);
and (ii) each operator here is fit on a capped $\le\!16{,}000$-pair/tick reservoir
($\zL\,0.58{\to}0.55$, $\zH\,0.20{\to}0.12$). The linear-vs-nonlinear gap
\emph{within} this table is unaffected.}
\label{tab:nonlinear}
\begin{tabular}{lccc}
\toprule
 & linear & kernel & MLP \\
\midrule
solved $\zL$            & $0.55$ & $0.56$ & $0.57$ \\
solved $\zH$            & $0.12$ & $0.14$ & $0.16$ \\
unsolved $\zL$          & $0.55$ & $0.56$ & $0.57$ \\
unsolved $\zH$          & $0.17$ & $0.18$ & $0.20$ \\
\midrule
scheduled-LDS (control) & $0.84$ & $0.84$ & $0.84$ \\
state-switch (positive) & $0.65$ & $0.90$ & $0.92$ \\
\bottomrule
\end{tabular}
\end{table}

\section{Step size and matched temporal granularity}
\label{app:stepsize}

A fair objection to the timescale split is that the two streams are not on the
same footing: $\zL$ takes four updates per segment under a drive held constant
within an H-cycle, $\zH$ two. A finer timestep buys higher one-step predictability
for purely kinematic reasons, so the split could be a coarse-graining artifact
rather than a statement about computation. Split-half correction de-biases
estimation noise and App.~\ref{app:injection} rules out the injection as the
predictor; neither addresses this.

The natural control is to plot per-tick $R^2$ against \emph{relative} step size
$\|\dz\|/\|z\|$ and ask whether the two streams fall on one curve. That control
is unavailable in this architecture, for a reason worth stating rather than
discovering twice: the blocks are post-norm with RMSNorm, so $\|z\|$ is pinned.
Measured across every recorded tick of both streams, the median state norm is
$22.63$ with a coefficient of variation of $0.000$. Relative step size is
therefore a constant rescaling of absolute step size---the rank correlation
between them is exactly $+1.000$---and normalising by $\|z\|$ cannot separate
anything.

What can still be said without it: on a single estimator (median per-tick $R^2$
over the solved stratum), restricting both streams to their overlapping
step-size range moves the gap from $+0.060$ to $+0.050$, i.e.\ the ordering does
not come from $\zL$ occupying a gentler part of the step-size axis.

\paragraph{The composition control, run.} The sharper control is to remove the
granularity difference outright: sample $\zL$ only at H-cycle boundaries, so
differencing yields one \emph{composed} update per (segment, H-cycle) instead of
$L$ one-step updates. The composed stream ($\zLc$) then occupies $\zH$'s clock
grid exactly---$30$ \emph{fittable} buckets each, of the $31$ ticks that grid
has---and every downstream step, split-half halves, high-energy weighting,
held-out scoring and the $0.3$ reliability floor, applies unchanged.
\emph{Fittable} is not \emph{reliable} in this paper's sense, and the gap between
the two is large enough to state: Algorithm~\ref{alg:fullclock} ranks those $30$
down to a top tercile of $10$, and the floor then filters $8$ of those out, so
each slow-grid stream reaches the aggregate with the $n{=}2$ the table reports.
Everywhere else here, ``reliable'' counts buckets that survive both steps.

\begin{center}\small
\begin{tabular}{@{}lccc@{}}
\toprule
stratum & $\zL$ (one-step) & $\zLc$ (composed) & $\zH$ \\
\midrule
solved   & $0.72$ \, ($n{=}11$) & $0.59$ \, ($n{=}2$) & $0.37$ \, ($n{=}2$) \\
unsolved & $0.80$ \, ($n{=}5$)  & $0.63$ \, ($n{=}2$) & $0.39$ \, ($n{=}2$) \\
\bottomrule
\end{tabular}
\end{center}

\noindent The split survives: at matched granularity the solved-stratum gap is
$+0.22$ ($0.59$ vs.\ $0.37$), so the fast loop is more linear than the slow one
for reasons that are not the clock. But the control is not free, and we state its
cost rather than only its verdict. Composition removes \textbf{$37\%$} of the
headline gap ($+0.35 \to +0.22$; $41\%$ on unsolved), so a substantial minority of
the reported split \emph{was} granularity. And the composed estimate inherits
$\zH$'s scarcity---$n{=}2$ reliable high-energy buckets, because fewer, larger
updates survive the reliability floor less often---so it is a two-bucket
comparison of the kind we flag elsewhere in this paper. The claim we take from it
is the ordering, not the size.

\paragraph{What a position-wise operator cannot represent.}
A second objection reaches further. Each operator is a single
$512\!\times\!512$ map fitted per clock tick on update vectors pooled over
puzzles \emph{and} positions, so it carries no term by which position $i$'s
update depends on position $j$'s state. The model, being a transformer, mixes
positions at every block through self-attention. A position-wise fit can absorb
cross-position effects to the extent that they are linearly predictable from
the local update; effects it cannot predict contribute to the residual.
Thus ``poorly linearisable'' cannot be read as ``nonlinear in its own state''.

Using the same estimator for both streams does not remove this limitation:
differences in the strength or local predictability of cross-position effects
can affect the fast--slow $R^2$ ordering. The association with constraint
propagation in Sec.~\ref{sec:compute} is compatible with such an explanation,
but does not distinguish it from other sources of prediction error.

Testing this explanation would require a cross-position predictor evaluated
under the same held-out protocol, checking whether the fast--slow $R^2$ gap
narrows. We did not fit one. Until it is
fitted, the supported statement is the one Sec.~\ref{sec:discussion} makes: low
held-out $R^2$ is limited prediction by the tested families, not a mechanism and
not irreducible nonlinearity.

\section{Sensitivity to energy and reliability thresholds}
\label{app:sensitivity}
Two hyperparameters were fixed by hand: a reliability-ceiling floor of $0.3$ and
a top-third energy cut. Both enter the same aggregate, so we sweep them---floor
$\in\{0.2,0.3,0.4\}$ $\times$ fraction $\in\{0.2,\tfrac13,0.5\}$ $\times$
$\{$energy-weighted, unweighted$\}$, $18$ cells per setting, on the solved
stratum. Nothing about the estimator changes; only what it is asked.

\begin{center}\small
\begin{tabular}{@{}lcccc@{}}
\toprule
setting & cells & gap min & median & flips \\
\midrule
HRM seed $0$      & $18$ & $+0.107$ & $+0.350$ & $0$ \\
HRM seed $1$      & $12$ & $+0.074$ & $+0.232$ & $0$ \\
composed $\zLc$   & $18$ & $+0.049$ & $+0.157$ & $0$ \\
TRM ($N{=}4096$)  & $18$ & $-0.143$ & $+0.080$ & $2$ \\
\bottomrule
\end{tabular}
\end{center}

\noindent (``gap'' is fast minus slow held-out hi-energy $R^2$; a cell is
unscorable when a threshold empties one side, which is why seed~$1$ has $12$.)

\paragraph{Reliable-tick coverage.}
At floor $0.2$ and top-half energy selection, the slow stream retains
$14$ ticks on seed 0 and $10$ on seed 1; the gaps remain $0.37$ and $0.26$.
Relaxing the floor alone increases the counts from $2$ to $9$ and from
$1$ to $5$, respectively. Sparse coverage under the default settings
therefore depends on both thresholds.

\paragraph{TRM sensitivity.}
Two of TRM's eighteen solved-stratum cells have negative gaps ($-0.02$ and
$-0.14$, both unweighted). This is consistent with the absence of a solved
split at $N=4096$ ($0.48/0.48$); this sweep does not test the reported
unsolved-stratum inversion.

\section{Scaled endpoint Jacobian spectra}
\label{app:jacobian}

\begin{figure}[t]
\centering
\includegraphics[width=\tabwidth]{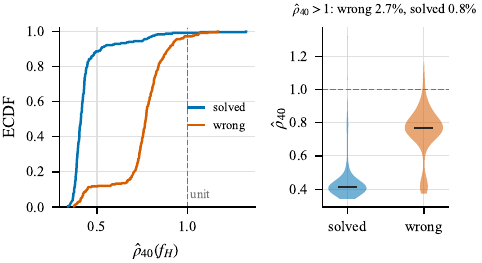}
\caption{Endpoint $f_H$ leading-Ritz radius, $256$ solved endpoints vs.\ $256$
wrong terminal states ($\mathrm{AUC}{=}0.913$). \textbf{Left:} ECDF, unit line
dashed. \textbf{Right:} violin by class; the tail past one is a distributional
property ($7/256$ wrong, $2/256$ solved), not one outlier. Per-example
complex-plane spectra: App.~Fig.~\ref{fig:jacobian}.}
\label{fig:jacobian_ecdf}
\end{figure}
\begin{figure*}[t]
\centering
\includegraphics[width=\linewidth]{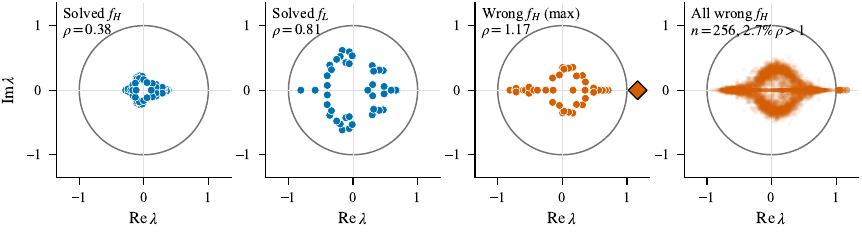}
\caption{Endpoint Jacobian Ritz values ($m=40$ Arnoldi; unit circle shown).
\textbf{Panels 1--2:} representative solved $f_H$ and $f_L$ spectra
($\hat\rho_{40}=0.38,0.81$).
\textbf{Panel 3:} the largest wrong-$f_H$ estimate ($1.17$).
\textbf{Panel 4:} all $256$ wrong-$f_H$ spectra. Estimates exceed one
for $7/256$ wrong and $2/256$ solved endpoints; after the terminal-update
gate, the counts are $4/176$ and $0/243$.
These are local linearisation estimates; a stability or saddle
interpretation additionally requires an appropriate fixed point.}

\label{fig:jacobian}
\end{figure*}

The Sec.~\ref{sec:endpoint} measurement draws $256$ solved and $256$ wrong
terminal states (wrong sampled proportionally across the $1$--$5$/$6$--$20$/$>\!20$
wrong-cell buckets, $85/85/86$), all through the same endpoint-Jacobian JVP $+\,m{=}40$
Arnoldi estimator; only per-example scalars are kept (no traces). Here and in
the legacy figure labels, $\rho$ abbreviates the estimated leading-Ritz radius
$\hat\rho_{40}$, not a certified full spectrum. Fig.~\ref{fig:jacobian} shows the
retained Ritz values themselves in the complex plane, which the main-text
distribution summarises away. The $n{=}16$
pilot behind its first three panels replicates at scale: solved $f_H$ contracts (mean $\rho{=}0.44$, gate-passing max
$0.82$, $65\%$ complex; the $3$ solved states above $0.9$ all fail the terminal-update gate),
wrong $f_H$ spreads higher (mean $0.75$, median $0.77$), with $9.4\%$ above $0.9$
and $2.7\%$ with an estimated expanding leading mode ($\hat\rho_{40}{>}1$; $7/256$, $95\%$ Wilson CI
$[1.3,5.5]\%$), so the expanding-mode estimate recurs across several examples
rather than one outlier (Fig.~\ref{fig:jacobian_ecdf}). Neither tail is
reported without the other. Raw, the two classes both have one: $2/256$ solved
endpoints exceed unity and the single largest radius in the whole dataset
($1.32$) is a solved state, above the wrong-class maximum of $1.17$. Gated to the
terminal-update criterion the picture is one-sided, because both solved outliers
fail that gate: $0/243$ gated solved endpoints exceed unity (max $0.82$) against
$4/176$ gated wrong ones (max $1.17$, itself a gate-passing endpoint). This is the
same gate the TRM comparison in App.~\ref{app:trm} applies, and the reason the
main text prefers $\mathrm{AUC}{=}0.913$ over either tail as the class summary.
Written once, the gate of Sec.~\ref{sec:setup} is
\begin{equation}
\begin{aligned}
\mathrm{gate}(i) &= \Big[\ \|\dz^{\mathrm{final}}_i\|\big/\|z_i\|\ \le\ q_{95}\ \Big],\\[2pt]
q_{95} &= \mathrm{pctl}_{95}\Big\{\|\dz^{\mathrm{final}}_j\|\big/\|z_j\|\Big\}_{j\ \mathrm{solved}} ,
\end{aligned}
\label{eq:gate}
\end{equation}
and every gated number in this paper---here, App.~\ref{app:maze},
App.~\ref{app:trm}, App.~\ref{app:relabel}---is this expression with the solved
class of its own setting supplying $q_{95}$. Calibrating on the solved class,
rather than on an absolute tolerance, is what lets one gate travel between
architectures and tasks whose state norms differ; it also means $95\%$ of solved
endpoints pass by construction, so the gate is only ever informative about the
wrong class. $69\%$ ($176/256$) of wrong endpoints clear it vs.\ $95\%$ of solved;
wrong-state estimated leading-Ritz radius correlates only weakly with failing that
gate ($r{=}0.22$). Within the wrong class that tail sits in the \emph{mildest} bucket
($5/85$ at $1$--$5$ wrong cells, versus $1/85$ at $6$--$20$ and $1/86$ at
${>}20$; max $\rho{=}1.17$)---the opposite of a severity-tracks-instability
reading, consistent with the small-step-but-wrong
mild-failure regime of Sec.~\ref{sec:fail}, and a further reason the $2.7\%$ is
reported as a distributional tail rather than as a mechanism.

\section{Replay-label consistency and weighted sampling}
\label{app:relabel}
Every stratified analysis here pairs a solved/wrong label taken from the
\emph{frozen evaluation dump} with a state the analysis \emph{replayed itself}.
Our reproducibility statement records that the instrumented loop matches the
training forward pass to \texttt{bfloat16} tolerance; what it did not record is how many
endpoints live inside that tolerance. We measured it, because an endpoint whose
replay decodes differently from the dump is analysed under a label that does not
describe it.

\paragraph{How often the two disagree.} Re-decoding each replayed endpoint and
comparing to the dump gives $467/512$ agreement overall, and the disagreement is
strongly asymmetric across severity:

\begin{center}\small
\fittab{%
\begin{tabular}{@{}lcccc@{}}
\toprule
dump stratum & agree (batch $1$) & $n$ & agree (batch $128$) & $n$ \\
\midrule
solved            & $0.953$ & $256$ & $0.957$ & $512$ \\
wrong $1$--$5$    & $\mathbf{0.722}$ & $90$ & $\mathbf{0.741}$ & $170$ \\
wrong $6$--$20$   & $0.919$ & $86$ & $0.947$ & $170$ \\
wrong ${>}20$     & $0.988$ & $80$ & $0.983$ & $172$ \\
\bottomrule
\end{tabular}}
\end{center}

\noindent The two columns are separate draws, not the same endpoints replayed
twice, so their agreement bounds how large a batch-composition effect could be
rather than excluding one. Within each, the effect is that endpoints one or two
cells from correct sit inside the \texttt{bfloat16} tolerance, and $27.8\%$ of
the mild bucket re-decodes as fully solved. Concordance is now emitted by every
basin run.

\paragraph{Re-indexing the rates by the state actually analysed.} Relabelling
moves $45$ of $512$ endpoints across the solved/wrong split and at least $70$
across the four severity buckets; the mild bucket loses more than half its
members---not only the $25$ that become solved, but others whose replay error
count exceeds five. The paper's contrast \emph{widens} under the more
self-consistent labelling:

\begin{center}\small
\fittab{%
\begin{tabular}{@{}lcccc@{}}
\toprule
& \multicolumn{2}{c}{dump label (this draw)} & \multicolumn{2}{c}{replay label} \\
\cmidrule(lr){2-3}\cmidrule(lr){4-5}
class & $R_{\mathrm{ans}}$ & $n$ & $R_{\mathrm{ans}}$ & $n$ \\
\midrule
solved          & $0.953\,[0.93,0.98]$ & $256$ & $0.993\,[0.98,1.00]$ & $277$ \\
wrong $1$--$5$  & $0.400\,[0.31,0.50]$ & $90$  & $0.325\,[0.18,0.47]$ & $40$  \\
wrong $6$--$20$ & $0.128\,[0.06,0.20]$ & $86$  & $0.041\,[0.01,0.08]$ & $98$  \\
wrong ${>}20$   & $0.025\,[0.00,0.06]$ & $80$  & $0.010\,[0.00,0.03]$ & $97$  \\
\bottomrule
\end{tabular}}
\end{center}

\paragraph{Reporting rule and sampling limitation.}
The relabelled groups inherit unequal inclusion probabilities from the
dump-stratified draw; their unweighted rates are conditional on that draw.
The mild group also shrinks to $n{=}40$, with a wide interval.
We retain the original dump-labelled headline for consistency with
TRM and Maze, and report replay labels as a sensitivity analysis.
The job script specified before execution that the headline would be retained
if the replay-mild upper confidence bound remained below $\tfrac12$ and
disjoint from the solved interval. This condition held.

This draw is separate from the larger replication in App.~\ref{app:eps0},
whose $512$ solved and $170$ mild-wrong endpoints supply the main-text rates.
Subsequent basin runs emit both labellings. The probability-sampling audit
below addresses the unequal-inclusion issue directly.

\paragraph{Sampling on the replay label, rather than relabelling after the draw.}
The objection above is a design defect, not a presentation one: no reordering of
an already-drawn sample fixes it. So we redrew. Every valid test example is
assigned its dump stratum, the $934$-example mild stratum is taken as a
\emph{census} and the other three at known rates
($4000/176{,}574$ solved, $2000/47{,}142$, $2000/198{,}136$), each drawn example
is replayed once at batch size $1$, and the sample is stratified on what
that replay \emph{decodes to}. Inclusion probabilities are known by construction,
and the class rates are inverse-probability weighted,
$\hat R_C{=}\sum_i w_i\mathbf{1}[C_i{=}C]r_i/\sum_i w_i\mathbf{1}[C_i{=}C]$ with
$w_i{=}1/(\pi_{1i}\pi_{2i})$. The draw has two stages: the $8934$-example screen
above is stage $1$, and stage $2$ takes the $1952$ endpoints the sweep perturbs,
$512$ per replay stratum except mild, where only $416$ were available and all
were taken. This is the H\'ajek ratio form, which corrects for the unequal
inclusion probabilities and is design-consistent for the frame; it is not
finite-sample unbiased, and we do not claim it is. Within a class $\pi_{2i}$ is
constant, so it cancels from $\hat R_C$ but not from any population total. Intervals resample endpoints within each design stratum. A uniform draw
was rejected on measurement: the design's own Horvitz--Thompson estimate puts the
replay-mild population at $845$ of $422{,}786$ examples ($0.200\%$; the dump-mild
stratum is $934$, $0.221\%$), so landing $n{=}170$ replay-mild that way needs
${\sim}85{,}000$ replays against this design's $8934$. Every endpoint re-decoded to the label the sampler had assigned
it ($1952/1952$), so the replay is deterministic and the probabilities describe
the endpoints they are attached to.

\begin{center}\small
\fittab{%
\begin{tabular}{@{}lcc@{}}
\toprule
class & dump-drawn (published) & replay-drawn, IPW/H\'ajek \\
\midrule
solved            & $0.953\,[0.934,0.971]$ & $1.000$ \\
wrong, $1$--$5$   & $0.418\,[0.347,0.494]$ & $\mathbf{0.178}\,[0.156,0.201]$ \\
wrong, $6$--$20$  & $0.088\,[0.047,0.135]$ & $0.020\,[0.009,0.033]$ \\
wrong, ${>}20$    & $0.041\,[0.017,0.076]$ & $0.064\,[0.043,0.087]$ \\
\bottomrule
\end{tabular}}
\end{center}

\noindent Rates are $R^{(C)}_{\mathrm{ans}}(0)$, with
$n=512/416/512/512$. The solved--mild separation widens: $3.5\%$ of
dump-solved endpoints replay as wrong, while $30\%$ of dump-mild endpoints
replay as solved, inflating the dump-mild persistence rate.
Three qualifications remain. The observed solved rate of $1.000$ has a
degenerate bootstrap and is not evidence of zero population uncertainty.
Severity is no longer monotone: the $6$--$20$ bucket lies below the $>20$
bucket on disjoint intervals. Mild remains the most persistent wrong bucket.
Finally, the headline retains dump labels for cross-setting consistency;
the replay-weighted audit establishes robustness of the separation, not
equivalence of the two estimands.

\paragraph{Does anything \emph{else} in the paper depend on the labelling?
A post-hoc sensitivity check.} The redraw above repairs the stationarity result,
which is the one the headline rests on. Two other results are label-dependent and
were not redrawn: Sec.~\ref{sec:endpoint}'s endpoint spectral separation and
Sec.~\ref{sec:fail}'s probe-versus-q-halt comparison, both computed on the
\emph{dump}-stratified $512{+}512$ sample. Redrawing them would mean re-running the
Arnoldi estimator on a new probability sample, which we did not do. Instead we
relabel the endpoints already measured, using the terminal H-lens decode the
instrumented forward pass had already recorded ($\hat y = \arg\max\,
\mathrm{lm\_head}(\zH)$ at the last H-step), so no new forward pass is needed.
\textbf{This is a post-hoc sensitivity analysis, not a replacement estimator}: the
sample was drawn stratified on the dump label, so a replay-label group is
whatever fell into it rather than a stratified sample of its class---the same
design objection this appendix raises against the relabel-after-the-draw table
above, and the reason the redraw exists. Nothing below changes a reported number.
On this draw the two labellings agree on $83.1\%$ of severity buckets and on
$92.4\%$ of the solved/wrong split, with the disagreement concentrated in the
same place as before ($95.7\%$ agreement on solved, $41.2\%$ on mild, of which
$25.9\%$ replay as fully solved).

\emph{The spectral separation does not depend on the labelling.} On the same
$512$ endpoints, the leading-Ritz AUC moves from $0.913\,[0.886,0.940]$ under the
dump label ($256/256$) to $0.928\,[0.902,0.953]$ under the replay label
($275$ solved / $237$ wrong); a paired bootstrap over the relabelled endpoints
puts the shift at $+0.014\,[-0.011,+0.039]$, an interval containing zero. Only
$37$ of $512$ endpoints move class ($28$ wrong$\to$solved, $9$ the other way), and
restricting to the $475$ on which the two labellings agree gives $0.950$. The
expanding tail is untouched in its numerators---the same $7$ wrong and $2$ solved
endpoints exceed unity under either label; only the denominators shift, so the
wrong-class rate reads $2.7\%$ or $3.0\%$. One detail \emph{is} label-sensitive
and we flag it: the strict one-sidedness after gating ($0/243$ solved vs.\
$4/176$ wrong) survives relabelling under the published gate ($0/257$ vs.\
$4/162$), but if $q_{95}$ in Eq.~\ref{eq:gate} is instead recomputed over the
\emph{replay}-solved residuals it becomes $1/261$ solved against $6/216$ wrong.
The tail stays roughly six times more common among wrong endpoints, but ``zero on
the solved side'' is a property of the published gate rather than a
label-independent fact.

\emph{The probe comparison survives, and strengthens.} Absolute AUROCs are not
comparable across the two labellings here, and the reason is worth stating rather
than working around: the replay label is a deterministic function of the same
terminal $\zH$ that both the probe and the q-halt head read, so relabelling
inflates \emph{both} scores towards $1$ (probe $0.993$, q-halt $0.995$ overall).
What the paper's claim rests on is the \emph{gap}, and the gap is what to compare.
Under the dump label the probe beats q-halt by $+0.006$ overall and $+0.012$ on
the mild bucket; under the replay label it does not beat it at all ($-0.002$ and
$-0.014$). Sec.~\ref{sec:fail}'s conclusion---that correctness is not
substantially more linearly decodable than the halting head already extracts---is
therefore understated as published, not overstated, on either labelling.
The dump-label column
reproduces Sec.~\ref{sec:fail}'s $0.941$/$0.887$ to within $0.002$ under a
different scikit-learn build; we quote the published values in the body.

\paragraph{Every setting, at its own published configuration.}
Everything above concerns Sudoku/HRM. The other five settings reported
dump-labelled rates with nothing to compare them against, so each published
sweep was re-run at its own checkpoint, sample, seed, endpoints-per-class and
wrong strata, with the radius list truncated to that setting's headline radius
plus the $\varepsilon{=}0$ control. The perturbation seed is a function of
(endpoint, radius, seed) alone, so the surviving radius is unchanged and the
dump column has to reproduce the published number. It does, in all five
settings that have a published counterpart. Seed $1$ was instead re-run at
seed $0$'s configuration, which makes the two seeds severity-comparable for the
first time and leaves it without such a check.

\begin{center}\small
\fittab{%
\begin{tabular}{@{}lcccccc@{}}
\toprule
& & \multicolumn{2}{c}{solved $R_{\mathrm{ans}}$} &
    \multicolumn{2}{c}{wrong $R_{\mathrm{ans}}$} & gap \\
\cmidrule(lr){3-4}\cmidrule(lr){5-6}
setting & concordance & dump & replay & dump & replay & change \\
\midrule
HRM $\cdot$ Sudoku, seed $0$ & $0.918$ & $0.954$ & $0.992$ & $0.181$ & $0.072$ & $+0.15$ \\
HRM $\cdot$ Sudoku, seed $1$ & $0.892$ & $0.918$ & $0.971$ & $0.147$ & $0.033$ & $+0.17$ \\
TRM $\cdot$ Sudoku           & $0.818$ & $0.958$ & $1.000$ & $0.323$ & $0.000$ & $+0.37$ \\
HRM $\cdot$ Maze             & $0.833$ & $0.599$ & $0.755$ & $0.301$ & $0.258$ & $+0.20$ \\
HRM $\cdot$ N-Queens $10{\times}10$ & $0.953$ & $0.958$ & $1.000$ & $0.559$ & $0.532$ & $+0.07$ \\
TRM${+}$GRAM $\cdot$ N-Queens & $0.969$ & $0.983$ & $0.967$ & $0.835$ & $0.847$ & $-0.03$ \\
\bottomrule
\end{tabular}}
\end{center}

\noindent Wrong-class rates pool the severity buckets the setting drew from.
Concordance is below $1$ everywhere and the disagreement is never symmetric.
Maze is the extreme: $28\%$ of its dump-solved endpoints replay as wrong,
which is most of what its weak published separation was. The solved-minus-wrong
gap widens in five of the six and narrows only for the GRAM-guided recurrence,
the setting with the highest concordance and so the least room to move. The
ordering solved $>$ wrong holds $6/6$ under either label. Dump labelling was
understating the separation rather than manufacturing it, which is the same
direction the Sudoku redraw above found.

\emph{What a replay bucket means on a multi-solution task.} The replay severity
code counts cells differing from the \emph{canonical} board, so on N-Queens a
replay-mild endpoint may be a different valid solution rather than a near-miss.
The solved/wrong split is unaffected---both labels mean ``exactly
canonical''---but the two N-Queens rows' replay \emph{buckets} are distance from
the canonical board, not severity of failure.

\emph{How exactly these reproduce.} The TRM re-run first missed its published
counterpart by exactly one endpoint per class, with a perturbation scale of
$14.13$ against the published $10.82$, on the same seed, sample and code. The
published run used an L40S and the re-run an RTX 6000 Ada; TRM does not contract
at the horizon ($\lVert\Delta\zH\rVert/\lVert\zH\rVert\approx0.11$), so a
different floating-point reduction order moves $z^\ast$ and with it the median
one-step displacement that sets the noise amplitude. Repeating the re-run on an
L40S returns $10.82$ and the published rates exactly, and the GRAM setting,
re-run on the same device as its published counterpart, reproduced its scale to
six digits throughout. The table reports the L40S run. Reproduction of any rate
here is therefore good to about one endpoint per class across devices, and we
do not claim more.

\paragraph{``The wrong class'' is not the same estimand twice.}
Each setting reports one wrong-class rate and the six are read side by side,
but each is a mixture over severity and the mixtures are set by whichever binds
first, the sampler's cap or the pool itself. In the test frame, mild failures
are $0.4\%$ of Sudoku's wrong endpoints ($934$ of $246{,}212$), $2.4\%$ of
seed $1$'s, $0.2\%$ of TRM's ($10$ of $6{,}591$) and $13.4\%$ of Maze's
($120$ of $893$); the two N-Queens settings have no mild wrong bucket at all,
because there stratum $1$ is a valid answer and the wrong class is drawn from
the invalid strata only. In the analysed samples the caps invert part of that:
mild is a third of Sudoku's wrong class and a quarter of Maze's, but $1\%$ of
TRM's and none of the GRAM run's, whose $64$ wrong endpoints are $95\%$
severe. So the settings' wrong-class levels average over compositions ranging
from a third mild to none, which is a second reason---independent of the
per-run $\kappa$ of Sec.~\ref{sec:discussion}---that the levels are not
comparable across settings while the within-setting ordering, measured on one
draw under one protocol, is.

\section{Pre-specified and exploratory analyses}
\label{app:prereg}
Thirteen of the $63$ job scripts carry a literal \texttt{PRE-REGISTERED READ}
tag, $14$ occurrences in all: $13$ distinct rules, since \texttt{rans\_interval}
and \texttt{rans\_severity} share one verbatim and one script carries two. A
fourteenth, the replay sampler, carries a separate \texttt{PRE-COMMITTED READ}
rule. These are internally \emph{pre-specified}, not externally
preregistered. The original endpoint and timescale analyses are exploratory;
the table distinguishes their follow-up tests from post-hoc checks.

\begingroup\small
\begin{audittable}{@{}p{0.24\linewidth}p{0.18\linewidth}p{0.44\linewidth}@{}}
\toprule
claim / analysis & status & the rule, and which branch fired \\
\midrule
\ificlr\endhead\fi
\multicolumn{3}{@{}l}{\emph{Confirmatory: a rule was fixed and committed before the run.}}\\[2pt]
Wrong answers are non-stationary (Sec.~\ref{sec:fail})
& pre-specified & Stands only if solved and mild intervals are disjoint and
mild's upper end stays below $\tfrac12$. $0.953$ vs $0.418\,[0.347,0.494]$: it did. \\
Replay-label audit (App.~\ref{app:relabel})
& pre-specified & Keep the dump-labelled headline if the separation survives. It
survived on the replay-first probability sample the rule names
($0.178\,[0.156,0.201]$ vs $1.000$); we kept it. \\
Horizon (App.~\ref{app:horizon})
& pre-specified, twice & The first read allowed ``wrong states do reach an
attractor, merely slowly''---the largest correction available to this paper. It
did not fire ($0.43$, no trend, through $2048$ segments). \\
Near-losslessness of the $\zH$ site (Sec.~\ref{sec:intervention})
& pre-specified, \emph{overtaken} & Separating Wilson intervals would leave the
claim standing; overlapping ones would retire it on power grounds. They
separated \emph{against} it ($\zH$ regresses $2.5\times$ as often). That branch
was not written down, and we retired the claim anyway. \\
Granularity control (App.~\ref{app:stepsize})
& pre-specified & Survives at gap ${\ge}0.20$, dies at ${\le}0.10$, report the
residual in between. It cleared by $0.02$ ($+0.22$), and we wrote it up as the
in-between branch. \\
TRM inversion (App.~\ref{app:trm})
& pre-specified & At $N{=}4096$: adequate slow-tick coverage would remove
the caveat. The slow cell remained $n{=}2$; the caveat remains. This does not
prove larger samples cannot improve coverage. \\
Noise vs.\ a task-specific baseline (App.~\ref{app:tta})
& pre-specified & Interval covering zero $\to$ noise adds nothing over symmetry.
It did not: $+0.51\,[+0.29,+0.73]$, $21/0$. The gain is real and small. \\
Maze to competence (App.~\ref{app:maze-clean})
& pre-specified & The original rule treated no improvement beyond $10.7\%$
as evidence against training length. The run had median $13\%$, peak $27\%$,
below the $40\%$ competence target; the causal inference is unsupported. \\
N-Queens null (App.~\ref{app:nqueens})
& pre-specified & The rule contrasted a data-regime null with task specificity.
Matched-size Sudoku reached $4.3\%$; cross-task differences prevent
isolating a task-specific cause. \\
Maze endpoint spectra (App.~\ref{app:maze})
& pre-specified & Third branch named in advance: a solved class too small to
decide $\to$ report underpowered, do not spend a second run. It fired, on all
$107$ solved mazes. \\
Seed-$1$ full-clock cell (Fig.~\ref{fig:coverage})
& pre-specified; later refit & The original run omitted the promised
reliable-bucket counts. A later stored refit reports $11$ fast and $1$ slow
bucket on solved states; the slow-loop estimate remains sparsely supported. \\
\addlinespace
\multicolumn{3}{@{}l}{\emph{Exploratory or post-hoc: no rule was fixed in advance.}}\\[2pt]
Endpoint Jacobian spectra (Sec.~\ref{sec:endpoint})
& exploratory & The original measurement pipeline; the confirmatory work above
tests what it found. \\
The timescale split itself (Sec.~\ref{sec:transient})
& exploratory & Discovered, not predicted. Its controls were pre-specified; the
finding was not. \\
Threshold sensitivity (App.~\ref{app:sensitivity})
& post-hoc & Run to answer the $n{=}1$--$3$ objection after it was raised. \\
Replay-label sensitivity of spectra and probe (App.~\ref{app:relabel})
& post-hoc & Asks whether anything \emph{other} than the stationarity result
moves under replay labels. The main comparisons persist; gating details change. \\
Injection timing (App.~\ref{app:timing})
& post-hoc & Most recoveries at injection index $4$ precede the q-halt plateau;
the proxy does not establish an escape mechanism. \\
Expanding-mode control (App.~\ref{app:expanding})
& post-hoc, rebuilt & An earlier version returned the opposite verdict; the
control families are now pinned by unit tests. \\
Endpoint robustness census
& exploratory & Sampling was cut by hand after the problem was seen; it backs no
claim here. \\
\bottomrule
\end{audittable}
\endgroup

\noindent Two departures require disclosure: the regression-rate comparison
returned an unanticipated direction, so the lower-regression claim for $z_H$
was withdrawn; the original seed-1 full-clock job did not record its promised
reliable-bucket counts. The later refit fills the artifact gap but does not
retroactively satisfy the original reporting rule.

\section{Compute-matched symmetry-augmentation baseline}
\label{app:tta}
Sudoku is invariant under a symmetry group (digit relabeling, band/stack and
within-band/stack row/column permutations, transpose): applying any composition to a
puzzle yields an equivalent instance whose solution is the transform of the original.
For each puzzle in the uniform $N{=}4096$ subset we draw $k{=}8$ seeded random
compositions, run the standard $16$-segment forward on each variant, inverse-transform
the predicted board, and reselect across the $9$ candidates by max final q-halt
(compute-matched to the $\zH$ injection). The transforms round-trip exactly and
preserve Sudoku validity (unit-tested).

\paragraph{What the noise adds on top.} The two arms run on the same
uniform subset, so their candidates can be pooled without extra inference: one
baseline, $8$ symmetry variants, $8$ noisy reruns, one max-q-halt argmax over all
$17$.

\begin{center}\small
\begin{tabular}{@{}lcc@{}}
\toprule
pool & $\Delta$ (pts) & gained / lost \\
\midrule
symmetry only ($k{+}1$)      & $+34.64$ & $1423$ / $4$ \\
noise only ($k{+}1$)         & $+4.76$  & $197$ / $2$ \\
combined ($2k{+}1$)          & $+35.16$ & $1444$ / $4$ \\
\midrule
\emph{marginal of noise}     & $\mathbf{+0.51}$ \, $[+0.29,+0.73]$ & $21$ / $\mathbf{0}$ \\
\bottomrule
\end{tabular}
\end{center}

\noindent The marginal is real---the paired interval excludes zero---and it is
$21$ puzzles. Of the $197$ the injection fixes, $176$ ($89\%$) are also fixed by
some symmetry variant. So against a task-specific baseline that subsumes it, the
recurrent-state intervention contributes about a tenth of its standalone $+4.8$.

On the $962$ puzzles outside the nine-orientation oracle union, noise recovers
$2.2\%$, all selected by q-halt. This subset uses ground-truth labels and is a
headroom diagnostic, not a deployable selection rule.

\textbf{Result.} Exact accuracy $0.416\to0.762$ ($+34.6$ points, $95\%$ CI
$[+33.2,+36.1]$; $1423$ gained, $4$ lost, $99.8\%$ retained). Per-cell majority vote
across the $9$ candidates does \emph{not} help ($-1.1$ points), so confidence-based selection performs better than this voting rule.

\textbf{Validity (not a leak).} Each orientation individually solves ${\sim}41\%$
(baseline $0.416$; the eight variants $0.405$--$0.415$), but different orientations
solve different puzzles: at least one of the nine is correct for $76.5\%$ of puzzles
(the oracle union). The q-halt-chosen accuracy $0.762$ sits below this ceiling and
recovers $99.6\%$ of it, and \emph{zero} q-halt-chosen predictions are correct outside
the union, an internal-consistency check on candidate selection. This check alone
does not exclude every possible source of label leakage.

\textbf{Overlap.} Symmetry augmentation and noise share $176$ recovered puzzles;
symmetry uniquely recovers $1247$. This demonstrates sensitivity to puzzle
orientation, but does not identify an attractor mechanism.

\section{Injection site, timing and rerun budget}
\label{app:timing}

\label{app:ablation}
All numbers below (Table~\ref{tab:ablation}) use the frozen intervention
configuration (noise scale $\varepsilon=3.0$, injection at zero-based ACT index~$4$,
max-$q$-halt reselection) on the disjoint uniform $N{=}4096$ subset; only labels
enter, as a score. They are a \emph{re-run} of that configuration under a
different candidate seed sequence: both runs seed the injection at
$\mathrm{seed}{+}1$ and differ only in the per-candidate stride, so the first of
the eight $\zH$ reruns repeats the same noise draw and the other seven are
fresh. That is why $\zH$ reads $+4.6$ ($191/2$)
here against Sec.~\ref{sec:intervention}'s $+4.8$ ($197/2$): the same puzzles
under the same procedure, a replication of the perturbation rather than of the
sample. The injection index $s_{\mathrm{inject}}$ is zero-based, unlike the
segment index $s\!\in\!\{1..16\}$ of Sec.~\ref{sec:setup}:
$s_{\mathrm{inject}}{=}4$ injects after five of the sixteen segments have
completed, and $s_{\mathrm{inject}}{=}0$ after one. The two recurrent sites are
perturbed at that point; the input embedding is perturbed once before the
recurrence starts and the perturbed embedding then drives every step, so the
input arm differs from the other two in timing and persistence as well as in
site.

\begin{table}[h]
\centering
\small
\caption{Reselection by injection site ($k=8$), using a common absolute
noise scale derived from the $z_H$ final update. This corresponds to
different relative magnitudes at the three sites.
The site-relative comparison below changes the input result from $+15.6$
to $-1.1$ points. The small regression counts here do not resolve a
difference between recurrent sites; the larger-frame comparison follows.}

\label{tab:ablation}
\fittab{%
\begin{tabular}{lcccc}
\toprule
site & $\Delta$ (pts) & $95\%$ CI & gained & lost \\
\midrule
$\zH$ (slow state)  & $+4.6$  & $[+4.0,+5.3]$   & $191$ & $2$  \\
$\zL$ (fast state)  & $+3.7$  & $[+3.1,+4.2]$   & $151$ & $1$  \\
input embedding     & $+15.6$ & $[+14.3,+16.9]$ & $714$ & $75$ \\
\bottomrule
\end{tabular}}
\end{table}

\paragraph{Regression rates on the full test frame.} Two lost against one, out of
$4096$, has no statistical content: at those rates the two sites are
indistinguishable below $N\!\sim\!10^5$, and the near-losslessness we claimed for
$\zH$ rested on exactly that comparison. Re-running the frozen $k{=}8$ recipe at
both recurrent sites on the \emph{full} $422{,}530$-puzzle test frame---nothing
was tuned on it, and the prediction frame replays exactly---makes the rates
estimable. ($422{,}530$ rather than the $422{,}786$ of
Sec.~\ref{sec:setup}: the $256$-puzzle pilot that fixed $\varepsilon$ and the
injection segment is held out, so no example that set a hyperparameter is scored
here.)

\begin{center}\small
\fittab{%
\begin{tabular}{@{}lccc@{}}
\toprule
site & $\Delta$ (pts) & gained / lost & regression rate, Wilson $95\%$ CI \\
\midrule
$\zH$ & $+4.36$ \, $[+4.30,+4.42]$ & $18{,}464$ / $38$ & $0.0090\%$ \, $[0.0066,0.0123]$ \\
$\zL$ & $+2.91$ \, $[+2.86,+2.96]$ & $12{,}313$ / $15$ & $\mathbf{0.0036\%}$ \, $[0.0022,0.0059]$ \\
\bottomrule
\end{tabular}}
\end{center}

\noindent The intervals separate, and in the direction opposite to the claim
they were run to test. $\zH$ regresses $2.5\times$ as often as $\zL$. So the
timescale split does not identify a \emph{safer} site; it identifies a
\emph{higher-gain} one, at a measurably higher regression rate. Among recurrent
sites there is a trade rather than a free lunch, and we have corrected the main
text accordingly. Two things survive: both recurrent sites remain two orders of
magnitude safer than the input embedding, which at matched magnitude is net
negative; and the $k$-sweep is monotone on the full frame as well ($+1.44$,
$+2.28$, $+3.26$, $+4.36$ for $k{=}1,2,4,8$), so extra compute still buys
accuracy rather than trading it.

The $k$-sweep at $\zH$ is monotonic: $\Delta=+1.6\,[+1.2,+2.0]$,
$+2.7\,[+2.2,+3.2]$, $+3.6\,[+3.1,+4.2]$, $+4.6\,[+4.0,+5.3]$,
$+5.8\,[+5.1,+6.5]$ points for $k=1,2,4,8,16$, with gained/lost counts
$68/1$, $110/1$, $151/2$, $191/2$, $239/2$. The regression count stays at
$1$--$2$ throughout, so extra compute buys accuracy without eroding the
high observed retention rate. Perturbing $\zL$ is nearly as effective ($+3.7$,
$151/1$): a $\zL$ kick propagates into $\zH$ within one H-cycle
through the coupling, so it acts as an \emph{indirect} $\zH$ perturbation rather
than an independent injection site. We treat the input-injection result cautiously:
although its net gain is larger, it perturbs the persistent input drive rather
than the recurrent state, testing a different hypothesis (input-space test-time
augmentation, the input-perturbation effect independently reported by
\citet{renliu2026}) than the escapability claim; its $75$ regressions (against $2$
for $\zH$) show it is not the low-regression, mechanism-aligned lever the
headline relies on. Per-cell \emph{majority} voting over the same $k{=}8$ reruns
is \emph{worse} than baseline ($-0.51$ points, $0.4106$ vs.\ $0.4158$), the figure
Sec.~\ref{sec:intervention} quotes: selection, not agreement, does the work. It is
not the $+0.51$ noise marginal of App.~\ref{app:tta}, a different quantity that
happens to move the same $21$ puzzles of $4096$.

\paragraph{Selection rule.}
On the same nine symmetry candidates, per-cell majority voting loses
$47$ net puzzles ($337$ gained / $384$ lost, $-1.15$ points), whereas
max-q-halt gains $34.6$ points ($1423/4$). Thus candidate diversity alone
does not explain the low observed regression rate.

\paragraph{Magnitude-matched sites: the input advantage does not survive.}
The comparison above perturbs every site at one common \emph{absolute} scale
($3.0\times$ the median $\|\dz_{\mathrm{final}}\|$), which is derived from the
$\zH$ update and is therefore not a comparable perturbation at the other two
sites. Re-running with each site scaled by its \emph{own} measured radius
($\zH$ by the norm of its final segment-to-segment update, $\zL$ by the norm of
its final L-step update, the input by the input-embedding norm; $N{=}4096$, all
else frozen) reverses the ordering. The two recurrent radii are measured over
different clock intervals, since with $H{=}L{=}2$ one ACT segment contains four
L-steps:

\begin{center}\small
\begin{tabular}{@{}lcc@{}}
\toprule
site & $\Delta$ (pts) & gained / lost \\
\midrule
$\zH$   & $+4.6$ & $191/2$ \\
$\zL$   & $+2.9$ & $121/1$ \\
input   & $\mathbf{-1.1}$ & $198/\mathbf{243}$ \\
\bottomrule
\end{tabular}
\end{center}

\noindent The $z_H$ scale is unchanged. Under site-relative scaling both
recurrent sites improve accuracy, whereas input injection loses $1.1$ points
($198$ gained / $243$ lost). Site rankings therefore depend on the scale
convention. State-update norms and input-embedding norms are different
reference quantities; this comparison does not establish a uniquely fair
cross-site perturbation metric.

\paragraph{Between sites, paired on the same puzzles.} Each delta above compares
one site to the shared baseline, which orders the sites only indirectly: two
intervals that do not overlap are not a test that one site beats another. Because
every rerun's prediction is retained, the sites can be compared directly. Holding
the puzzles, the budget ($k{=}8$) and the site-relative scaling fixed, and scoring
the max-q-halt choice at each site against the same labels, McNemar on the paired
outcomes gives

\begin{center}\small
\begin{tabular}{@{}lccc@{}}
\toprule
pair & $\Delta$ (pts) & $95\%$ CI & gained / lost \\
\midrule
$\zH$ vs.\ $\zL$    & $+1.68$ & $[+1.20,+2.17]$ & $86/17$ \\
$\zH$ vs.\ input    & $+5.71$ & $[+4.60,+6.83]$ & $396/162$ \\
$\zL$ vs.\ input    & $+4.03$ & $[+2.95,+5.11]$ & $342/177$ \\
\bottomrule
\end{tabular}
\end{center}

\noindent None of the three intervals covers zero, so at matched radii the
ordering $\zH > \zL >$ input is a paired result and not an inference from three
separate baselines. The gap between the two recurrent sites is small ($1.7$
points) beside either one's margin over the input embedding, which is the
comparison Fig.~\ref{fig:overview}(a) is drawn to make. This is a within-model
ranking at one frozen configuration; it does not carry to TRM, where injecting
into either loop gives the same lift (App.~\ref{app:trm}).

\paragraph{Why the input site fails: the candidates are fine, the ranking is not.}
Retaining every rerun's prediction lets us separate the two things a
reselection procedure does---generate candidates, and pick one. Write the
\emph{oracle union} for the fraction of puzzles some rerun gets right, and ask
how much of that reachable headroom max-q-halt actually captures:

\begin{center}\small
\begin{tabular}{@{}lccccc@{}}
\toprule
site & $k$ & max-q-halt & oracle union & headroom recovered \\
\midrule
$\zH$   & $8$  & $0.4619$ & $0.4624$ & $\mathbf{99.0\%}$ \\
$\zH$   & $16$ & $0.4736$ & $0.4741$ & $\mathbf{99.2\%}$ \\
$\zL$   & $8$  & $0.4451$ & $0.4453$ & $99.2\%$ \\
input   & $8$  & $0.4048$ & $\mathbf{0.4839}$ & $\mathbf{-16.1\%}$ \\
\bottomrule
\end{tabular}
\end{center}

\noindent Baseline accuracy is $0.4158$ ($N=4096$, site-relative scaling).
The final column is the fraction of \emph{oracle headroom} recovered,
$(a_{\mathrm{selected}}-a_{\mathrm{base}})/
(a_{\mathrm{oracle}}-a_{\mathrm{base}})$, not selected accuracy divided by
oracle accuracy. At recurrent sites q-halt recovers about $99\%$ of
available improvement. Input injection has the largest oracle union among
these three sites, yet selection yields a net loss. This identifies ranking
as a limitation at the input site; a distribution shift in the states scored
by q-halt is a possible explanation, not a measured causal mechanism.

\paragraph{Protocol details.}
Intervention hyperparameters (rerun count $k=8$, injection after five ACT segments
(zero-based index $4$), noise scale $\varepsilon=3.0\times\text{median}\|\dz_\text{final}\|$, and
selection by maximum final $q$-halt logit) were tuned on a disjoint
$256$-example pilot and then \emph{frozen} before the reported evaluation on the
uniform $N=4096$ subset, so no tuning on the reported evaluation subset enters the $+4.76$-point result.
The selection rule was chosen by comparison against per-cell majority voting and
a $q$-halt gap rule on the pilot; max-$q$-halt was retained based on its pilot performance. Relative ranking across reruns and
absolute correctness discrimination are evaluated separately. The stratified per-bucket gains
(bucket $1$--$5$ $+56.5\%$, $6$--$20$ $+22.9\%$, $>\!20$ $+2.9\%$) are much larger
than the uniform $+4.76$ points because the stratified sample deliberately
oversamples failures; the uniform subset gives the unbiased estimate. Only $2$
puzzles were lost against $197$ gained, and $99.9\%$ of originally-solved puzzles
were retained; regressions are defined relative to the baseline-solved set.

Two mechanisms could explain the Sec.~\ref{sec:intervention} gain: escape from a
wrong state whose trajectory has already committed, or ensemble diversity from
perturbing a trajectory that has not. To separate them we sweep the injection segment
$s_{\mathrm{inject}}\in\{0,2,4,6,8,12\}$ with the frozen recipe ($k{=}8$,
$\varepsilon{=}3.0\times$, max-q-halt) on the uniform $N{=}4096$ subset, holding
the injected noise \emph{fixed} across settings so that only the timing varies.
For each setting we also record the fraction of gained puzzles whose baseline
q-halt trajectory had already plateaued (first reached $95\%$ of its peak rise
above segment $0$) before the injection point. Read this only as a proxy for a \emph{late, committed}
trajectory, not as evidence of stationarity---Sec.~\ref{sec:fail} measures that
directly, and finds most wrong endpoints fail it---so recovery there is the more
escape-like of the two mechanisms rather than redirected exploration.

Earlier injection produces larger gains: $9.94$ points at
$s_{\mathrm{inject}}{=}0$ versus $1.10$ at $s_{\mathrm{inject}}{=}12$
(Table~\ref{tab:timing}).
At the frozen injection index $4$, $27\%$ of recovered puzzles had
reached the q-halt plateau before injection; at $s_{\mathrm{inject}}{=}12$,
$67\%$ had.
Most recovery at injection index $4$ therefore occurs before this plateau marker.
This is consistent with trajectory diversification, but does not partition
the gain causally into exploration and escape. Earlier injection also
allows more continuation, and the same absolute kick can have a different
relative scale. The q-halt plateau is not a stationarity certificate.

\begin{figure}[h]
\centering
\includegraphics[width=\tabwidth]{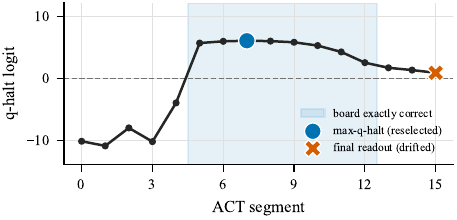}
\caption{\textbf{Why reselection works (one rescued puzzle).} The q-halt logit is
deeply negative until the board is exactly correct (shaded), then saturates; the
forced final segment has drifted back to wrong, the reselected max-q-halt segment
has not.}
\label{fig:rescue}
\end{figure}

\begin{table}[h]
\centering\small
\caption{Injection-timing sweep (uniform $N{=}4096$; noise fixed across rows).
$\Delta$ exact accuracy with paired McNemar $95\%$ CI, puzzles gained/lost, and
the fraction of gained puzzles whose baseline q-halt had already plateaued before
injection (a late-trajectory proxy, not a stationarity test).}
\label{tab:timing}
\fittab{%
\begin{tabular}{ccccc}
\toprule
$s_{\mathrm{inject}}$ & $\Delta$ (pts) & $95\%$ CI & gained/lost & plateaued \\
\midrule
$0$  & $+9.94$ & $[+9.01,+10.87]$ & $412/5$ & $0\%$  \\
$2$  & $+6.47$ & $[+5.71,+7.23]$  & $266/1$ & $6\%$  \\
$4$  & $+4.76$ & $[+4.10,+5.42]$  & $197/2$ & $27\%$ \\
$6$  & $+4.17$ & $[+3.56,+4.79]$  & $172/1$ & $38\%$ \\
$8$  & $+3.05$ & $[+2.52,+3.59]$  & $127/2$ & $56\%$ \\
$12$ & $+1.10$ & $[+0.78,+1.42]$  & $45/0$  & $67\%$ \\
\bottomrule
\end{tabular}}
\end{table}

\begin{figure}[h]
\centering
\includegraphics[width=\tabwidth]{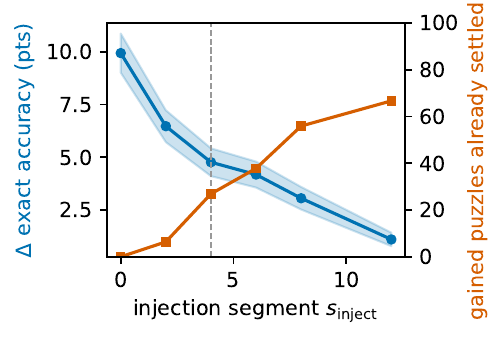}
\caption{Injection-timing sweep. Blue: $\Delta$ exact accuracy (paired $95\%$ CI
band) falls monotonically as the zero-based injection index $s_{\mathrm{inject}}$
(axis label ``injection segment'') moves later; the dashed line marks the
main-text $s_{\mathrm{inject}}{=}4$. Orange: the fraction of gained
puzzles whose trajectory had already plateaued (q-halt plateau---a late-trajectory
proxy, not a stationarity test) before injection rises with $s_{\mathrm{inject}}$,
so the timing sweep characterises both recovery and its relation to this
plateau marker, without identifying an escape mechanism.}
\label{fig:timing}
\end{figure}

\section{Compute-matched extended-depth baseline}
\label{app:depth}
The $k=8$ procedure costs $9\times$ inference compute. The depth baseline
uses the same budget for a single trajectory of $144$ ACT segments. Because the recurrence is a prefix of
itself, one $144$-segment pass yields every shorter rung, so the whole ladder is a
single sweep ($N{=}4096$ uniform, seed-0 checkpoint, no perturbation anywhere).

The baselines are matched in total recurrent segment count, not wall-clock
latency: extended depth executes $144$ sequential segments, whereas
perturb-and-reselect executes nine $16$-segment trajectories whose independence
permits concurrent evaluation. With sufficient accelerator capacity, concurrency
can reduce response time at the cost of greater simultaneous activation memory;
our experiments do not establish a measured parallel speedup. Extending the depth
also requires no extrapolation of a learned clock encoding. The exogenous clock
denotes the prescribed update schedule, and the segment index is not supplied to
the recurrent modules. The extended-depth baseline repeats the same weight-shared
updates with the unchanged $H=2,L=2$ schedule, carrying both latent states forward;
positional encodings remain tied to puzzle positions. No segment-index clipping,
modulo operation, or state reset is applied at the training horizon. Nevertheless,
execution beyond the $16$ training segments can encounter unfamiliar latent
states. We therefore treat the $144$-segment run as a horizon-extrapolation
baseline: its reported accuracy measures the utility of continued recurrence at
matched computation, without assuming unchanged representation quality or
equivalent deployment latency.

The injection arm in this comparison is the frozen recipe ($+4.8$), not the
sweep maximum of $+9.94$ at $s_{\mathrm{inject}}{=}0$ (App.~\ref{app:timing}),
which is matched on budget too. We compare against the frozen arm because its
$(\varepsilon,s_{\mathrm{inject}})$ was fixed on a disjoint pilot before this
evaluation, whereas the sweep maximum is selected post-hoc over six values on
the evaluation subset itself and carries the corresponding optimism. Read as a
ceiling rather than a protocol, well-timed injection exceeds extended depth at
the same budget.

\begin{center}\small
\begin{tabular}{@{}lccc@{}}
\toprule
segments & $\Delta$ (pts) & $95\%$ CI & gained / lost \\
\midrule
\multicolumn{4}{@{}l}{\emph{read out at the final segment}} \\
$144$ & $+5.6$ & $[+4.9,+6.4]$ & $245/14$ \\
\midrule
\multicolumn{4}{@{}l}{\emph{max-q-halt segment of the same single pass}} \\
$144$ & $\mathbf{+6.0}$ & $[+5.3,+6.7]$ & $\mathbf{248/2}$ \\
\midrule
\multicolumn{4}{@{}l}{\emph{oracle: any segment of the pass correct}} \\
$144$ & $+6.1$ & $[+5.4,+6.8]$ & $249/0$ \\
\bottomrule
\end{tabular}
\end{center}

\noindent This is a genuine head-to-head: the ladder draws the \emph{same}
uniform subset as the injection arm (valid examples only, tuning pilot excluded,
same seed), and the baselines agree to one puzzle---$0.41602$ against $0.41577$.

At the matched $9\times$ budget, final-segment depth gains $5.6$ points and
depth with segment selection gains $6.0$, versus $4.8$ for the frozen noise
recipe. Segment selection recovers $248/249$ oracle-recoverable puzzles and
reduces regressions from $14$ to $2$. These are observed comparisons on this
sample, not a proof of oracle-optimal selection.
The $245$ wrong-to-right changes under final-segment readout show that some
terminal errors can be corrected by continued recurrence; they do not
exclude later convergence to other attractors.

\section{Unperturbed continuation control}
\label{app:eps0}

The $\varepsilon=0$ control continues each terminal state without a kick.
It distinguishes ordinary continuation from perturbation-induced displacement.
The initial full-trajectory sample contains $32$ solved and $32$ mild-wrong
endpoints, with $12$ noise seeds and scale $\kappa=2.391$; the deterministic
control is run once per endpoint.

\begin{center}\small
\fittab{%
\begin{tabular}{@{}lcccc@{}}
\toprule
$\varepsilon$ & solved $R_{\mathrm{state}}$ & wrong $R_{\mathrm{state}}$ & median $d/d_0$ (solved) & median $d/d_0$ (wrong) \\
\midrule
$0$ (no perturbation) & $0.8125$ & $0.1250$ & $1.000$ & $1.000$ \\
$1.0\times$           & $0.8125$ & $0.1250$ & $1.004$ & $1.000$ \\
$19.95\times$         & $0.8099$ & $0.1250$ & $1.008$ & $1.000$ \\
\bottomrule
\end{tabular}}
\end{center}

\noindent Here $d_0$ is the displacement under unperturbed continuation.
At the largest radius the median solved displacement is $1.008d_0$;
the wrong-class ratio remains $1.000$. The mild-wrong answer-persistence
rate is $0.375$ without perturbation, consistent with the perturbed sweep.
Thus continuation dominates the tested perturbation effect. This pilot
uses the same severity bucket as the main text, but fewer endpoints.

\paragraph{Re-measured at $16\times$ the endpoints.} Because the pilot showed little sensitivity to radius, we allocated
more computation to endpoints---$\varepsilon{=}0$ plus one perturbed radius, at
$512$ endpoints per class---and report percentile bootstraps clustered on
endpoints, since the seeds of one endpoint share its geometry:

\begin{center}\small
\fittab{%
\begin{tabular}{@{}lccc@{}}
\toprule
class & $R_{\mathrm{ans}}$ & $R_{\mathrm{state}}$ & $n$ \\
\midrule
solved             & $0.953\,[0.934,0.971]$ & $0.856\,[0.824,0.885]$ & $512$ \\
wrong, $1$--$5$    & $0.418\,[0.347,0.494]$ & $0.129\,[0.082,0.182]$ & $170$ \\
wrong, $6$--$20$   & $0.088\,[0.047,0.135]$ & $0.047\,[0.018,0.082]$ & $170$ \\
wrong, ${>}20$     & $0.041\,[0.017,0.076]$ & $0.035\,[0.012,0.064]$ & $172$ \\
\midrule
wrong, pooled      & $0.182$                & $0.070$                & $512$ \\
\bottomrule
\end{tabular}}
\end{center}

\noindent The pilot mild rate lies inside the larger sample's interval.
Solved and mild-wrong intervals are disjoint, and the mild upper bound is
$0.494<\tfrac12$. Mild failures are the most persistent wrong bucket.
This ordering does not distinguish boundary sensitivity from continued state
motion. In the separate replay audit (App.~\ref{app:relabel}), $27.8\%$ of
dump-labelled mild endpoints re-decode as solved. Error count does not measure
distance to a decode boundary, and becoming fully solved on replay is a
different event from preserving the entire decoded answer under continuation.
The observed severity ordering therefore does not exclude boundary sensitivity
as a contributor to answer non-persistence.
The pooled $0.182$ is a mixture over the sampled severities, not a
test-population-weighted wrong-answer rate. The severity ordering changes
under replay labels (App.~\ref{app:relabel}).

These results support answer non-persistence at the tested horizon.
They do not measure basin depth, prove noise has no effect at other scales,
or identify the mechanism of mid-trajectory reselection.

\paragraph{It replicates on a second seed and a second architecture.}
Unperturbed continuation returns the perturbed rates again on the seed-1
checkpoint ($0.81/0.09$ against a perturbed median $0.81/0.10$; median $d/d_0
\le 1.05$) and on TRM ($0.23/0.03$ against $0.22/0.04$; $d/d_0 \le 1.09$). So
the perturbation failing to matter is a property of these recurrences, not of one
checkpoint. (TRM's absolute rates here differ from the $0.32/0.02$ quoted in
App.~\ref{app:trm}, which was measured on a different stratified sample; the
$\varepsilon{=}0$ vs.\ $\varepsilon{>}0$ comparison is within-run and unaffected.)

The perturbation sweep is therefore a robustness check on the continuation
measurement. Both class separation and its interpretation should be assessed
within each checkpoint and sampling protocol.

\section{Continuation to 128 times the trained budget}
\label{app:horizon}
The unperturbed control removes the perturbation as the cause of the motion. It
does not remove the \emph{horizon}. Our stationarity test continues for one more
budget ($16$ segments), so it establishes that a wrong terminal state is not
stationary \emph{at that horizon}, and a state can lie inside a basin without
having arrived. A reader is entitled to ask whether these states simply converge
later, to a wrong answer. We therefore continued them, with no perturbation, to
$512$ segments: $32\times$ the trained budget, reading the state out at
$16/32/64/128/256/512$. Solved states were run identically as a control.
This and the $2048$-segment run below are horizon extrapolations on the same
terms as App.~\ref{app:depth}: no segment index enters the recurrent modules, so
nothing is clipped or wrapped, but the latent states reached late in these runs
are further outside anything training visited than at $144$ segments. Drift
measured there may reflect that distance rather than the dynamics the trained
regime would show.

\begin{center}\small
\begin{tabular}{@{}lcccccc@{}}
\toprule
$T$ & $16$ & $32$ & $64$ & $128$ & $256$ & $512$ \\
\midrule
\multicolumn{7}{@{}l}{\emph{answer unchanged since the previous read-out}} \\
solved & $0.92$ & $0.97$ & $0.97$ & $0.97$ & $0.97$ & $\mathbf{0.97}$ \\
wrong  & $0.16$ & $0.33$ & $0.34$ & $0.42$ & $0.44$ & $\mathbf{0.45}$ \\
\addlinespace
\multicolumn{7}{@{}l}{\emph{answer unchanged since the terminal state}} \\
solved & $0.92$ & $0.92$ & $0.92$ & $0.92$ & $0.92$ & $0.92$ \\
wrong  & $0.16$ & $0.16$ & $0.19$ & $0.19$ & $0.17$ & $0.16$ \\
\addlinespace
\multicolumn{7}{@{}l}{\emph{exact-correct}} \\
wrong  & $0.23$ & $0.23$ & $0.25$ & $0.28$ & $0.30$ & $0.30$ \\
\bottomrule
\end{tabular}
\end{center}

\noindent The pilot uses $n=64$ per class. Wrong-answer persistence remains
below solved persistence, but successive-readout rates cannot be compared
directly because their intervals differ. We therefore extend continuation to
$2048$ segments with equal $128$-segment windows ($n=64$ per class;
wrong endpoints pooled over all three severity strata).
Selected checkpoints are shown below; each rate compares to the readout
$128$ segments earlier, including checkpoints not printed in the table.

\begin{center}\small
\fittab{%
\begin{tabular}{@{}lccccccc@{}}
\toprule
$T$ & $128$ & $256$ & $512$ & $1024$ & $1536$ & $2048$ & mean $256$--$2048$ \\
\midrule
wrong  & $0.19$ & $0.44$ & $0.44$ & $0.44$ & $0.45$ & $0.41$ & $\mathbf{0.43}$ \\
solved & $0.92$ & $0.97$ & $0.97$ & $0.97$ & $0.97$ & $0.97$ & $0.97$ \\
\bottomrule
\end{tabular}}
\end{center}

\noindent Across the fifteen windows after the first, wrong-answer persistence
ranges from $0.39$ to $0.45$ (mean $0.43$, fitted slope $+0.0002$ per window),
versus $0.97$ for solved endpoints. Wrong-class exact accuracy stabilises
around $0.28$--$0.30$. There is no detectable increasing persistence over
this measured horizon.

A pre-specified control expected the solved-state residual to decay toward
zero. The recorded quantity does not: it remains about $0.99$ of its
first-readout value through $512$ segments. That quantity is not the one-step
residual the control named. It is the displacement between successive readouts,
$\|\zH(T)-\zH(T_{\mathrm{prev}})\|$, normalised by the first readout's class
median, over windows that grow through $16/16/32/64/128/256$ segments. A
displacement measured over windows of changing length does not fix a
per-segment rate, so we read no convergence verdict from it. Complex eigenvalues alone also do not
establish a limit cycle. We therefore do not infer a rotating attractor from
this observation.

Finite continuation cannot exclude slower convergence, periodic dynamics,
or an attractor reached beyond $2048$ segments. The supported conclusion is
persistent answer changes over the observed windows, not the absence of
asymptotic attractors.

\section{Nonlinear correctness readouts}
\label{app:nonlinear-readout}

Sec.~\ref{sec:fail} reports that a \emph{linear} probe on the final $\zH$ barely
outperforms the model's own halting head, and we drew no design conclusion from
that, because testing one probe family licenses none. Fitting the missing
families settles it. Features are a \emph{trajectory summary} the $\zH$ probe
never sees---the per-segment halt and continue logits and the four sublayer
update norms at every recorded tick ($800$ features)---on the same $1024$-state
stratified sample, stratified $5$-fold cross-validated, scored out-of-fold, with
bootstrap CIs over states:

\begin{center}\small
\fittab{%
\begin{tabular}{@{}lcc@{}}
\toprule
readout & AUROC (all) & AUROC (mild only) \\
\midrule
q-halt (the model's own) & $0.934$ \, $[0.916,0.951]$ & $0.873$ \, $[0.835,0.909]$ \\
linear, trajectory       & $0.956$ \, $[0.943,0.968]$ & $0.919$ \, $[0.891,0.943]$ \\
MLP $(64,32)$            & $0.966$ \, $[0.956,0.976]$ & $0.942$ \, $[0.923,0.959]$ \\
gradient-boosted trees   & $\mathbf{0.969}$ \, $[0.958,0.979]$ & $\mathbf{0.944}$ \, $[0.921,0.962]$ \\
\bottomrule
\end{tabular}}
\end{center}

\noindent Both feature choice and model class contribute.
A linear trajectory readout improves mild-failure AUROC from the terminal-state
probe's $0.887$ to $0.919$; boosted trees on those same trajectory features
reach $0.944$. The latter exceeds q-halt's $0.873$, with non-overlapping
reported intervals. This is evidence of recoverable correctness information
beyond the tested terminal-state linear probe, not an attribution solely to
nonlinearity. The detector uses signals already produced by the model,
although recording them and fitting the readout incur additional cost.
Its across-puzzle discrimination is distinct from q-halt's performance
ranking reruns of one puzzle.

Decodability also emerges gradually over the iteration, rising from AUROC
$0.61$ at iteration 0 to $0.91$ mid-trajectory and $0.94$ at the end, mirroring
the contraction toward the low-residual endpoint.

\emph{Scope, and an artifact gap we should state.} $n{=}1024$ from the
stratified sample, one checkpoint, one task; no hyperparameter search was
performed, so these nonlinear models should be read as untuned reference
families rather than optimized predictors. The three fitted rows were re-derived after the
fact and now have stored metrics; the original fit was run outside our analysis
pipeline and its outputs were never kept, so feature construction and folds were
reconstructed from this description rather than recovered. Read them as an
independent re-derivation. The ordering and the q-halt comparison are unchanged
by it; the values moved by at most $0.011$. The $800$ trajectory features still
come from a residual-stream pass we do not persist (Reproducibility Statement),
so reproducing them means re-running from the checkpoint. The q-halt row is
quoted from the relabelling audit of App.~\ref{app:relabel}, which stores it;
the same logit captured inside the re-derivation scores $0.935/0.857$, and the
audit's $0.873$ sits inside that run's mild interval $[0.813,0.897]$.

\section{Three robustness checks: Krylov dimension, training maturity, second seed}
\label{app:robust}

\paragraph{Is the spectral separation an artifact of the Krylov dimension?}
Every radius we compute in this paper is a leading-Ritz estimate from an $m$-step
Arnoldi factorization with $m{=}40$, apart from the two sensitivity runs in this
paragraph; the third-party GRAM pilot of App.~\ref{app:gram} reports
$\hat\rho_{80}$. Re-running the endpoint sweep at $m{=}20$ and $m{=}80$ leaves the
class separation intact and moves the wrong-class maximum by less than
$10^{-7}$:

\begin{center}\small
\begin{tabular}{@{}lcccc@{}}
\toprule
$m$ & solved $\bar\rho$ & wrong $\bar\rho$ & wrong max & $n$/class \\
\midrule
$20$          & $0.419$ & $0.754$ & $1.166$ & $128$ \\
$40$ (paper)  & $0.442$ & $0.749$ & $1.166$ & $256$ \\
$80$          & $0.427$ & $0.757$ & $1.166$ & $128$ \\
\bottomrule
\end{tabular}
\end{center}

\noindent The $\rho{=}1.17$ endpoint is therefore not a subspace-truncation
artifact: doubling and halving the Krylov dimension moves it in the eighth
decimal ($1.1661161$ at $m{=}40$ and $m{=}80$, $1.1661162$ at $m{=}20$).
\emph{Caveat:} the $m{=}20$ and $m{=}80$ runs used $128$ endpoints per class
against the main run's $256$, so the \emph{tail counts} across rows are not a
matched comparison and we do not read the solved-side counts ($0/128$, $0/128$
against $2/256$) as evidence about the solved outliers.

\paragraph{Does the timescale split decay with training maturity?}
A reasonable worry is that the fast/slow dissociation is a late-training artifact
of a fully-trained network. It is not; it is present at all four training steps we
refit (held-out hi-energy $R^2$). The four step rows share one labelled manifest
built from the final seed-0 checkpoint, so each is fitted on the fixed cohort of
puzzles the final checkpoint solves rather than on a checkpoint-specific solved
stratum. The seed-1 row uses seed~1's own labels:

\begin{center}\small
\begin{tabular}{@{}lccc@{}}
\toprule
step & fast $\zL$ & slow $\zH$ & gap \\
\midrule
$15{,}624$ & $0.73$ & $0.47$ & $+0.26$ \\
$26{,}040$ & $0.87$ & $0.37$ & $+0.50$ \\
$36{,}456$ & $0.78$ & $0.33$ & $+0.45$ \\
$52{,}080$ (paper) & $0.72$ & $0.37$ & $+0.35$ \\
\addlinespace
seed $1$, $52{,}080$ & $0.61$ & $0.31$ & $+0.30$ \\
\bottomrule
\end{tabular}
\end{center}

\noindent The gap is large and positive throughout, and non-monotone rather than
decaying, so the split is a property of the trained recurrence and not of the
particular stopping point we analyse. The last row is a different \emph{seed}
rather than a different step: refitting the whole clock on seed $1$'s own sample
returns $+0.30$, so the split survives a change of initialisation as well as of
stopping point. That row carries the caveat of its own $n$, discussed next.

\paragraph{Second seed.} A seed-1 retrain (final exact accuracy $48.67\%$)
reproduces the intervention, more strongly than seed~0: $\zH$ gives $+6.6$ points
($274$ gained, $2$ lost), $\zL$ $+6.0$ ($252/6$), input $+16.6$ ($697/19$). These
numbers are scored against seed~1's \emph{own} predictions and are sound. We no
longer read the $2$-vs-$6$ split as evidence that $\zH$ is the safer site: at
$N{=}4096$ those counts are not separable, and on the full frame the ordering runs
the other way (App.~\ref{app:ablation}).

The \emph{stratified} geometry replicates too, and more sharply than on seed~0.
The two $R$ rows are the $n{=}32$-per-class full-trajectory set of
App.~\ref{app:eps0}, which is why they read $0.97/0.38$ rather than the
$0.95/0.42$ of Sec.~\ref{sec:fail}'s $512/170$ frame; the comparison is
seed-to-seed within a row.

\begin{center}\small
\fittab{%
\begin{tabular}{@{}lcc@{}}
\toprule
 & seed~0 & seed~1 \\
\midrule
endpoint $\bar\rho$ (solved / wrong) & $0.44$ / $0.75$ & $0.44$ / $\mathbf{0.85}$ \\
$\hat\rho_{40}{>}1$ (solved / wrong) & $2/256$ / $7/256$ & $3/128$ / $\mathbf{27/128}$ \\
gate-passing (solved / wrong)       & $0.95$ / $0.69$ & $0.95$ / $\mathbf{0.49}$ \\
$R_{\mathrm{state}}$ (solved / wrong) & $0.81$ / $0.12$ & $0.81$ / $\mathbf{0.09}$ \\
$R_{\mathrm{ans}}$ (solved / wrong)   & $0.97$ / $0.38$ & $0.88$ / $\mathbf{0.16}$ \\
\bottomrule
\end{tabular}}
\end{center}

\noindent Every direction the paper claims survives the seed change, and the
$\hat\rho_{40}{>}1$ tail---which we report two-sided throughout---is far more
asymmetric here ($21\%$ of wrong against $2\%$ of solved) than on seed~0. The
$\varepsilon{=}0$ result replicates as well: unperturbed continuation gives
$0.81/0.09$ against a perturbed median of $0.81/0.10$, with median $d/d_0$ at
most $1.05$.

\paragraph{Checkpoint-specific strata.}
Seed-1 analyses use labels from seed 1's own predictions. Reusing seed-0
strata mixed solved and wrong endpoints and produced misleadingly similar
answer-persistence rates ($0.74/0.72$). Rebuilding the sample restored the
contrasts above. Each replication therefore requires its own labelled manifest.

\section{N-Queens 12-by-12: low-accuracy training and a null intervention}
\label{app:nqueens}

Because every headline number in this paper is Sudoku, we pre-specified a third
task chosen to stress a property Sudoku lacks: N-Queens $12{\times}12$ has
$14{,}200$ complete boards in all, and each given-constrained puzzle admits a
subset of them, so a correct model need not reproduce the labelled one. The plan,
fixed in advance, set the board size, the scoring (a board is \emph{valid} if it
places exactly $12$ non-attacking queens and keeps every given queen;
\emph{canonical} if it equals the labelled solution), and the read: if both accuracies came out low, the verdict was underfit
and the run should go to the compute cap rather than be reported as a finding
about dynamics.

That is what happened. Trained to the cap on a single RTX~6000 Ada
($15$h$25$m), the model reaches $0.22\%$ canonical exact and $1.78\%$ \emph{any}
valid board ($1.56\%$ valid-but-non-canonical). A stratified draw for the failure
analysis found only \textbf{2} canonically-solved puzzles in the pool, against
$14$ valid-non-canonical, $17$ near-misses and $481$ far-misses, so there is no
solved class to compare a wrong class against, and none of the three analyses in
this paper can be run. The reselection arm is correspondingly null: over $900$
multi-solution puzzles, $k{=}8$ gives $\Delta_{\text{any}}{=}{+}0.22$ points
($95\%$ CI $[-0.21,+0.66]$) and solution-set coverage moves $0.50\%\to0.73\%$
against a deterministic ceiling of $28.6\%$.

The null bounds the task coverage: this checkpoint provides too few
canonical successes for the planned class comparisons. Differences in task,
scoring and data generation prevent attributing the result to augmentation
multiplicity alone.

\paragraph{A data-matched control, and what it does and does not explain.} To
separate the data regime from the task we rebuilt Sudoku at N-Queens' budget
($5{,}000$ puzzles with $5$ augmentations each, $30{,}000$ rows in $5{,}000$ groups) and trained the
byte-identical released recipe. Test exact accuracy, scored on the full
$422{,}786$-puzzle frame:

\begin{center}\small
\begin{tabular}{@{}lcccc@{}}
\toprule
epoch & exact & cell acc. & test loss & $q$-halt acc. \\
\midrule
$2000$ & $\mathbf{4.26\%}$ & $0.617$ & $1.46$ & $0.996$ \\
$4000$ & $3.45\%$ & $0.607$ & $1.78$ & $0.986$ \\
$6000$ & $3.61\%$ & $0.606$ & $1.91$ & $0.961$ \\
\bottomrule
\end{tabular}
\end{center}

\noindent Training exact accuracy reaches $0.99$--$1.00$, while test accuracy
remains $3.45$--$4.26\%$. This indicates a generalisation gap under the
smaller Sudoku data regime. The run ended at approximately $39\%$ of the
schedule, so later improvement cannot be excluded.
The matched-size Sudoku result shows that low accuracy is not inevitable at
this example count, but cross-task differences in solution multiplicity and
scoring prevent isolating a task-specific cause for the N-Queens null.

\section{Maze replication across training checkpoints}
\label{app:maze}
We re-run the headline analyses on a second weight-tied task,
Maze-$30{\times}30$-Hard (sequence length $900$, vocabulary $6$), under the
\emph{identical} released configuration ($H{=}2$, $L{=}2$, ACT max $16$). Our
single-GPU cluster cannot reach the paper's multi-GPU Maze-Hard accuracy, so we
analyse two points on one $\mathrm{gbs}{=}32$ trajectory: epoch $4000$
(uninterrupted; $1.5\%$ exact) and a warm-restart continuation to epoch $7000$
(re-warmed LR; $10.7\%$ exact). Both are under-trained, so this is a
reduced-scale check of whether the \emph{dynamics} transfer, not a performance
replication. Protocols are unchanged; the timescale test leans on the reliable
unsolved stratum ($512$ mazes, $R^2_{\mathrm{corr}}{\approx}1$), the solved
strata being thin ($15$ mazes at epoch $4000$). Stored metrics now cover
the spectra, persistence, full-clock fits and intervention results
(Tables~\ref{tab:maze-basin},~\ref{tab:maze-fullclock},~\ref{tab:maze-interv}).

\begin{table}[t]
\centering \small
\caption{Maze full-clock \emph{held-out} hi-E $R^2$ (Algorithm~\ref{alg:fullclock},
as Table~\ref{tab:fullclock}). \textbf{The $\Delta$ row is the result:} between
the two points $\zH$ does not move and the whole gap collapse is $\zL$ becoming
\emph{less} linearisable. Sudoku: unsolved $0.80$/$0.39$ ($+0.41$), solved
$0.72$/$0.37$ ($+0.35$). $\dagger$ epoch-$4000$ solved is $15$ mazes ($\zH$
$R^2_{\mathrm{corr}}{=}0.29$); read the \emph{unsolved} block.}
\label{tab:maze-fullclock}
\fittab{%
\begin{tabular}{llccc}
\toprule
stratum & training point & $\zL$ & $\zH$ & gap \\
\midrule
\multirow{3}{*}{unsolved}
 & epoch $4000$ ($1.5\%$)  & $0.94$           & $0.57$          & $+0.37$ \\
 & epoch $7000$ ($10.7\%$) & $0.61$           & $0.57$          & $+0.04$ \\
 & $\Delta$                & $\mathbf{-0.33}$ & $\mathbf{0.00}$ & $\mathbf{-0.33}$ \\
\addlinespace
\multirow{2}{*}{solved}
 & epoch $4000^\dagger$    & $0.90$           & $0.21^\dagger$  & $+0.69$ \\
 & epoch $7000$            & $0.59$           & $0.46$          & $+0.13$ \\
\bottomrule
\end{tabular}}
\end{table}

\begin{table}[t]
\centering \small
\caption{Maze reproduction at both levels (Sec.~\ref{sec:fail}; epoch $7000$;
$96{+}96$ states, $8$ seeds, isotropic $\zH$ noise at $5$ radii $\times$ median
$\|\dz_{\mathrm{final}}\|$; $768$ continuations per class per radius). Both levels order
the classes and both are flat in $\varepsilon$---the scale-invariance the main
text reports for Sudoku, and the only reason five radii are shown. The last
column is Sudoku-HRM on its $n{=}32$-per-class set, printed beside rather than in
prose for what it exposes: \textbf{every Maze \emph{solved} $R_{\mathrm{state}}$
cell sits below Sudoku's \emph{wrong} $0.12$}, so no absolute threshold separates
the classes across settings, while the within-setting ordering holds in all of
them.}
\label{tab:maze-basin}
\fittab{%
\begin{tabular}{llccccc@{\quad}c}
\toprule
 & class & $0.1$ & $0.32$ & $1.0$ & $3.2$ & $10$ & \emph{Sudoku} \\
\midrule
\multirow{2}{*}{$R_{\mathrm{state}}$}
 & solved   & $0.095$ & $0.095$ & $0.096$ & $0.098$ & $0.079$ & \emph{0.81} \\
 & wrong-FP & $0.027$ & $0.027$ & $0.026$ & $0.025$ & $0.022$ & \emph{\textbf{0.12}} \\
\addlinespace
\multirow{2}{*}{$R_{\mathrm{ans}}$}
 & solved   & $0.599$ & $0.583$ & $0.600$ & $0.600$ & $0.590$ & \emph{0.97} \\
 & wrong    & $0.301$ & $0.301$ & $0.276$ & $0.260$ & $0.275$ & \emph{0.38} \\
\bottomrule
\end{tabular}}
\end{table}

\paragraph{Endpoint spectra do not decide here, and say something else instead.}
We ran the endpoint Jacobians under the usual protocol on \emph{all} $107$ solved
mazes in the test set and $128$ wrong ones ($m{=}40$). The ordering is
\textbf{undecided}: $\hat\rho$ averages $1.003$ solved against $1.023$ wrong,
$+0.020$ with SE $0.024$ ($95\%$ CI $[-0.027,+0.066]$; medians $0.960$ vs.\
$0.966$), and it is not monotone in wrongness---the mildest bucket averages
$0.998$, \emph{below} solved. The Sudoku gap is $0.307$, fifteen times larger on
comparable $n$. We report this as underpowered rather than as a refutation, the
branch specified in advance for a solved class this small, and mark the cell in
Fig.~\ref{fig:coverage}. The solved side is a \emph{census}: at $10.7\%$ exact
there are only $107$ solved mazes, so this power cannot be bought with a larger
draw, only with a better checkpoint.

What the same numbers do establish is not about the ordering. Both classes sit at
$\hat\rho\approx1.0$, so here a \emph{solved} maze is not the stable rotating
contraction a solved Sudoku is ($0.44$); it is marginal. With the answer-level
rates above---solved mazes reproduce their own answer $0.60$ of the time against
Sudoku's $0.97$---two independent measurements agree that this network's correct
answers are not settled states. That is consistent with settling being something
a recurrence acquires with competence (App.~\ref{app:nqueens}), and is why the
low \emph{absolute} rates in Table~\ref{tab:maze-basin} are the expected reading
rather than a failure of the criterion. Which of the two causes---different task,
or under-training---produces the marginality, one such checkpoint cannot say.

\begin{table}[t]
\centering \small
\caption{Maze $\zH$-noise reselection (epoch $7000$; $k{=}4$,
$s_{\mathrm{inject}}{=}4$, $\varepsilon{=}3.0\times$, max-q-halt; $N{=}1000$, the
full test set). Baseline is the paired instrumented re-run. Sudoku at $k{=}4$:
$+3.6$ $[+3.1,+4.2]$ (App.~\ref{app:ablation}); the $k{=}8$ headline is $+4.76$.}
\label{tab:maze-interv}
\begin{tabular}{ccccc}
\toprule
baseline & chosen & $\Delta$ (pts) & $95\%$ CI & gained/lost \\
\midrule
$0.099$ & $0.134$ & $+3.5$ & $[+2.15,+4.85]$ & $42/7$ \\
\bottomrule
\end{tabular}
\end{table}

\paragraph{Attenuation is the fast loop moving, not the split dissolving.}
The split reproduces at epoch $4000$---a $+0.37$ gap against Sudoku's $+0.41$,
both fits reliable---and the later checkpoint has a gap of $+0.04$. The $\Delta$
row of Table~\ref{tab:maze-fullclock} says how: the slow loop is
\emph{unchanged}, and the whole collapse is the fast loop becoming less
linearisable. The model therefore becomes \emph{less} captured by the tested
linear families with more training, the opposite of the training-phase artifact
that the bare gap collapse would otherwise suggest. What we cannot separate is
``more training'' from the LR re-warm, which returned to peak and could itself
perturb the dynamics.

\paragraph{The intervention transfers; its accounting does not.}
$\zH$-noise reselection lifts exact accuracy $+3.5$ points
(Table~\ref{tab:maze-interv}), essentially Sudoku's $k{=}4$ figure ($+3.6$). The
regressions are what does not transfer, and the raw counts understate it because
the frames differ in size: $42/7$ on $N{=}1000$ against $151/2$ on $N{=}4096$ is
a loss \emph{rate} of $0.70\%$ against $0.049\%$, $14\times$ higher. The $0.099$
baseline is the instrumented loop's score, ${\sim}0.8$ points below the
training-eval $10.7\%$---the loop-vs-loop difference verified to
\texttt{bfloat16} tolerance in the reproducibility statement.

\textbf{Net.} Split, intervention and the solved-vs-wrong basin ordering all
appear on a second task. A \emph{fully-trained} Maze is what would settle the
three things this checkpoint leaves open: whether the split persists under
unperturbed training, whether the endpoint ordering exists, and whether the
marginal spectra are the task or the under-training.
App.~\ref{app:maze-clean} reports training without a warm restart and its
full-clock refit. The split persists, but the runs also differ in training
length and accuracy, so the cause of the magnitude change remains unresolved.

\section{Maze training without a warm restart}
\label{app:maze-clean}
To test whether the warm restart explained the reported $10.7\%$ Maze
accuracy, we ran the same configuration without restarting optimisation.
Resume sidecars preserve optimizer moments, step count, data-loader position
and provenance; resuming without a sidecar is rejected. Two $23$-hour
single-GPU jobs reached epoch $13{,}500$ ($421{,}875$ steps), versus the
original $7000$ epochs. The second job resumed from step $406{,}250$.

The recipe uses \texttt{lr\_min\_ratio=1.0}: after $2000$ warmup steps,
the logged learning rate stays at $10^{-4}$. A warm restart therefore
repeats warmup and resets Adam moments, rather than restarting a decaying
learning-rate schedule.

Across $25$ evaluations from epoch $1500$ to $13{,}500$, exact accuracy
ranges from $0.3\%$ to $27.0\%$, with median $13.0\%$ and no clear upward
trend. The original $10.7\%$ is within this range and recurs at epoch
$11{,}000$. Cell accuracy varies less ($0.949$--$0.978$); whole-maze
correctness magnifies these differences across $900$ cells.
This run does not support warm restart as the sole explanation for low
accuracy, but it is not a controlled estimate of its causal effect.

The pre-specified $40\%$ accuracy target was not reached. Maze therefore
remains a reduced-scale replication.

\paragraph{The full-clock refit this run was missing.} Refitting the final
no-restart checkpoint (step $421{,}875$, $23.0\%$ exact on the $1000$-maze test
split) gives held-out hi-energy $R^2$ of $0.856$ for $\zL$ against $0.465$ for
$\zH$ on solved states, a gap of $+0.391$, and $0.857$ against $0.500$ on
unsolved ($+0.357$). The published checkpoint reads $+0.129$ and $+0.038$. The
split therefore also appears in the no-restart run, with the same sign and a
larger observed magnitude. This comparison does not identify the causal effect
of the restart on the gap.

Three qualifications travel with that. The two runs differ in three ways at once,
restart, length and accuracy, so this is not a controlled estimate of the
restart's effect on the gap any more than the accuracy comparison above was.
The $23.0\%$ is one draw from a band this run swings across without trend, so it
is not evidence of a better-trained Maze. And the gaps do not order with accuracy
across the three Maze checkpoints we have:

\begin{center}\small
\begin{tabular}{@{}lccc@{}}
\toprule
checkpoint & exact & solved $\zL$ / $\zH$ & gap \\
\midrule
step $125{,}000$ & $1.5\%$  & $0.901$ / $0.212$ & $+0.689$ \\
step $218{,}750$ (published) & $10.7\%$ & $0.590$ / $0.462$ & $+0.129$ \\
step $421{,}875$ (no restart) & $23.0\%$ & $0.856$ / $0.465$ & $+0.391$ \\
\bottomrule
\end{tabular}
\end{center}

\noindent Steps $125{,}000$ and $218{,}750$ are one lineage either side of the warm
restart and step $421{,}875$ is a separate run, so these three points span two
optimisation histories rather than one, and the non-monotonicity is not a trend. What survives is the sign, positive
in all three, on the $11$/$5$ and $7$/$2$ reliable high-energy buckets the clock
affords here. The attenuation Sec.~\ref{sec:transient} reports is a property of the
published checkpoint, not of the task.

\section{GRAM endpoint pilot and GRAM-guided TRM}
\label{app:gram}
Maze (App.~\ref{app:maze}) changes the task and holds the architecture fixed;
this appendix and App.~\ref{app:trm} change the \emph{architecture}. Here it is GRAM, the
stochastic latent recurrence of \citet{gram2026} (not the unrelated
architectures sharing that acronym). This leg is deliberately small and is
reported for one reason: it produced a methodological warning that does not
depend on its sample size, and that we then had to apply to TRM.

Read the two halves separately. The protocol comparison is a \emph{population}
measurement over the full held-out splits ($n{=}773$ and $n{=}578$); the spectral
half is a balanced $n{=}2$-per-cell pilot that establishes wiring, not effect
sizes.

\paragraph{The sampler is load-bearing (Table~\ref{tab:gram-protocol}).}
Collapsing the stochastic transition to its conditional mean costs $49.8$ points
on N-Queens ($0.964\!\to\!0.466$) and $71.5$ on edge-colouring. Whatever the
latent recurrence is doing, sampling is part of it.

\begin{table}[h]
\centering \small
\caption{Solve rate of the \emph{same} checkpoint under three noise protocols,
prescanned over the whole held-out split (not the $n{=}2$ spectral strata, so
these rates are population-level). Sampled inference reproduces the
independently recorded single-draw scores ($0.952$ and $0.670$), a harness
check; collapsing the sampler to its conditional mean is catastrophic.}
\label{tab:gram-protocol}
\begin{tabular}{lcc}
\toprule
noise protocol & N-Queens & edge-col. \\
                          & ($n{=}773$) & ($n{=}578$) \\
\midrule
conditional mean ($\xi{=}0$) & $0.466$          & $0.005$ \\
quenched $T$-periodic        & $0.696$          & $0.291$ \\
sampled inference            & $\mathbf{0.964}$ & $\mathbf{0.720}$ \\
\bottomrule
\end{tabular}
\end{table}

\paragraph{No ordering claim, in either direction.}
A stochastic map has no single endpoint Jacobian, so the endpoint protocol has to
be read through a forcing choice (Table~\ref{tab:gram-endpoint}). Three of the
four cells interleave the validity strata outright. The fourth orders them the
\emph{wrong} way and rests on a valid stratum drawn from $3$ eligible items in
$578$, with $2$ batch-vs-single replay mismatches among $129$ candidates checked.
At $n{=}2$ per cell one cannot detect an ordering that exists, so we claim none.
What the sample does support is the weaker and here sufficient statement that a
contracting endpoint is not by itself evidence of correctness, which
Sec.~\ref{sec:endpoint} reaches for HRM from the opposite side, a wrong terminal
state there carrying a perfectly contracting local linearisation.

\begin{table}[h]
\centering \small
\setlength{\tabcolsep}{4pt}
\caption{Endpoint leading-Ritz estimates (median of $n{=}2$ per cell) for the
bare conditional $f_H$ map, the conditional $f_L$ map, and the joint
supervision-step map, with the joint map's fixed-point residual. The last column
states the $f_H$ verdict for the valid/invalid pair it heads, read off the two raw
per-item radii rather than the medians: ``none'' means the strata's radii
interleave, so the pair carries no ordering. Only one of the four pairs orders,
and it orders the \emph{wrong} way (v${>}$i: valid $\{1.09,1.34\}$ above invalid
$\{0.86,1.07\}$) in the one block where nothing is near a root.
$\dagger$ marks cells whose residual makes the radius an endpoint linearisation
only, not a stability claim; the $12.85$ entry is the median of $\{0.95, 24.75\}$
and the $24.75$ state has both the run's largest residual ($0.61$) and its largest
$m$-to-$m$ spread ($2.9{\times}10^{-2}$).}
\label{tab:gram-endpoint}
\begin{tabular}{llccccl}
\toprule
 & & \multicolumn{3}{c}{$\hat\rho_{80}$} & joint & $f_H$ \\
\cmidrule(lr){3-5}
protocol & stratum & $f_H$ & $f_L$ & joint & residual & order \\
\midrule
\multicolumn{7}{l}{\emph{N-Queens}} \\
mean     & valid   & $0.79$ & $0.69$ & $0.94$ & $2{\times}10^{-2}$ & none \\
mean     & invalid & $0.80$ & $1.46$ & $12.85^\dagger$ & $3{\times}10^{-1}$ & \\
quenched & valid   & $0.75$ & $0.47$ & $\mathbf{0.23}$ & $3{\times}10^{-5}$ & none \\
quenched & invalid & $0.76$ & $0.78$ & $\mathbf{0.38}$ & $5{\times}10^{-5}$ & \\
\midrule
\multicolumn{7}{l}{\emph{edge-colouring}} \\
mean     & valid   & $1.21^\dagger$ & $0.88$ & $1.87^\dagger$ & $3{\times}10^{-1}$ & v${>}$i$^\dagger$ \\
mean     & invalid & $0.97^\dagger$ & $1.15$ & $1.01^\dagger$ & $3{\times}10^{-1}$ & \\
quenched & valid   & $0.85$ & $0.61$ & $\mathbf{0.75}$ & $2{\times}10^{-3}$ & none \\
quenched & invalid & $0.80$ & $1.37$ & $1.05^\dagger$ & $6{\times}10^{-2}$ & \\
\bottomrule
\end{tabular}
\end{table}

\paragraph{Residuals and forcing protocol.}
Under quenched forcing on N-Queens, the $16$-step stroboscopic map has
relative residuals $3.2$--$8.7\times10^{-5}$ on all four tested states and
leading-Ritz estimates in $[0.19,0.57]$, for both valid and invalid answers.
Under the conditional-mean protocol, residuals reach $0.61$ and one
radius estimate reaches $24.75$. These are different deterministic maps.
Large-residual endpoints cannot support a fixed-point stability
interpretation; small residuals and finite-Krylov spectra are supporting
diagnostics, not certificates. The TRM analysis likewise reports its
spectra both before and after residual-based gating
(App.~\ref{app:trm}).

\begin{figure*}[t]
\centering
\includegraphics[width=\linewidth]{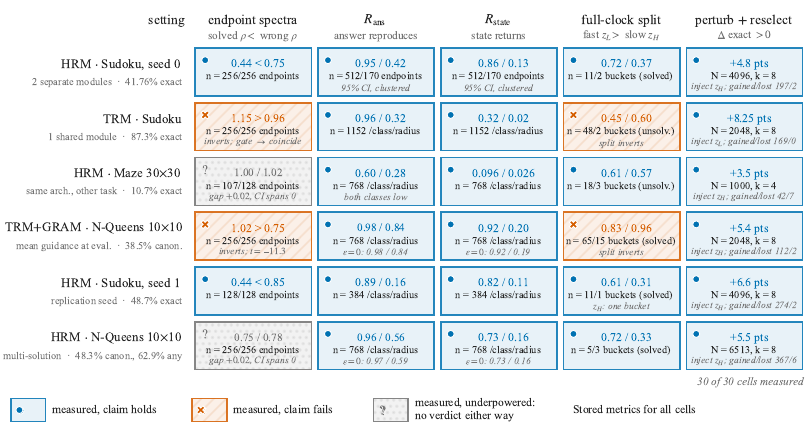}
\caption{\textbf{Coverage across six evaluated settings.}
Columns report endpoint spectral ordering, the full-clock split, state
persistence, answer persistence and reselection gain. Discs indicate the
specified comparison holds, crosses that it fails, and question marks that
the measurement is inconclusive. Each cell reports its sample or tick count;
all cells now have stored metrics.
Persistence columns denote solved--wrong ordering, not a universal
$\tfrac12$ threshold. That threshold fails for N-Queens and GRAM-guided TRM.
The TRM spectral inversion is ungated and disappears after gating;
its full-clock inversion occurs only in the unsolved stratum.
N-Queens-$12\times12$ and the third-party GRAM endpoint pilot are reported
separately in Apps.~\ref{app:nqueens} and~\ref{app:gram}.}

\label{fig:coverage}
\end{figure*}

\paragraph{Scope of the pilot.} $2$ items per
(protocol\,$\times$\,validity) cell, hence $8$ endpoint records per task; effect-size estimation requires a larger, purpose-designed sample. Nothing in the main text rests on these
numbers, and the edge-colouring results appear in Fig.~\ref{fig:coverage} not at
all and only here. As in the main text $\hat\rho_{80}$ is a finite-Krylov estimate
over a $69{,}632$- or $81{,}920$-dimensional state, not a certified radius. Unlike
the HRM and TRM legs this one runs against a third-party checkpoint through a
GRAM-side script that is not ours: we record the checkpoint, dataset and
analysis-code hashes with the metrics, but that leg cannot be re-run from our own
materials alone. That is why it stays a pilot, and why the figure's row
is no longer built from it.

\paragraph{GRAM-guided TRM.}
We train a reproducible GRAM-guided TRM on N-Queens-$10\times10$ with
\texttt{gram\_guidance=true}, $\beta=0.005$, guidance on $z_H$,
$H=3$, $L=6$ and $16$ ACT segments. It reaches $38.5\%$ canonical exact
accuracy and $50.4\%$ any-valid accuracy. At evaluation guidance is the
deterministic update $z_H\leftarrow z_H+\mu_p(z_H)$.
The original persistence sweep omitted $\varepsilon=0$. A separate control
on the same checkpoint gives answer persistence $0.98/0.84$, close to the
perturbed rates below, supporting a continuation-based interpretation.

Answer persistence is $0.98/0.84$ for solved/invalid boards, and state
persistence is $0.92/0.20$ ($768$ continuations per class per radius).
Reselection gains $5.4$ points ($95\%$ CI $[4.4,6.4]$, $112/2$,
$N=2048$). The raw endpoint-radius ordering reverses
($1.02$ solved versus $0.75$ wrong; $t=-11.3$, $256$ endpoints per class),
as does the full-clock split ($R^2=0.83$ fast versus $0.96$ slow).
The additive guidance update is a possible contributor to the high slow-loop
predictability; its causal contribution has not been isolated.

The terminal-update gate passes $94.9\%$ of solved and $96.5\%$ of wrong
endpoints. Together with the high wrong-answer persistence, this limits
generalisations of the HRM failure pattern. It also demonstrates that
high terminal persistence can coexist with gains from an earlier
intervention: the two experiments measure different behaviours.

\paragraph{Forward-pass parity.}
Earlier analysis replicas omitted the guidance step at the end of each
segment. This produced invalid but plausible metrics, including an endpoint
ordering of the opposite sign. Those artifacts are retained as superseded
and are not used here. All five analysis paths now call the same recurrence
function. The parity check covers the intervention rollout and the endpoint
rollout, precedes the analysis chain, and gives cell agreement $1.000000$. It
does not cover the instrumented full-clock driver, which carries the guidance
shift forward but records each $z_H$ snapshot at the output of
\texttt{L\_level}. The guidance update is therefore folded into the following
tick rather than given a tick of its own, and all fifteen high-energy $z_H$
buckets behind the $0.96$ above carry that folded tick on one side of their
regression.

\section{Relation to prior accounts of failure}
\label{app:disagreement}
The accounts reviewed in Sec.~\ref{sec:related} motivate a distinction
between within-budget trajectory geometry and persistence beyond the terminal
readout. \citet{ptrm2026} discuss bounded non-convergence;
\citet{corbett2026} discuss continued exploration and provide local
tube-stability and guide-alignment diagnostics.
These accounts need not describe mutually exclusive behaviours.

\paragraph{Source scope.}
The full text of \citet{efstathiou2026} is behind an access challenge we could
not clear, so everything we attribute to it comes from its published abstract,
which states all three of the claims at issue: that ``failed runs plateau at
stable high-loss attractors rather than continuing exploration'', that
trajectories ``descend into local minima determined by initialization'', and
that the intervention is ``adding noise to answer initialization and scaling
guess attempts''. We make no claim to have verified the experiments behind
them, and we do not describe those experiments. The distinction our
measurement turns on survives either way: changing an initial state can select
a different basin even when attractors are stable, so a successful
initial-state perturbation does not test terminal-state stability.

\paragraph{Different stability quantities.}
The tube-stability result of \citet{corbett2026} assumes contraction along a
neighbourhood of the deterministic trajectory and estimates local secant
ratios for its diagnostics. Our Arnoldi calculation estimates leading
eigenvalues at selected endpoints. A spectral-radius estimate is not a
uniform operator-norm bound, and a small terminal residual does not certify
a fixed point. Our measurements therefore neither verify that theorem's
assumptions nor establish their failure on the checkpoints used there.
Instead, Sec.~\ref{sec:fail} directly measures whether a terminal decoded
answer survives a specified continuation, with an unperturbed control.
Its finite horizon limits conclusions about asymptotic attraction.

\paragraph{Candidate generation and selection.}
Both prior studies distinguish available candidates from the selector's
ability to recover them. On Maze, \citet{ptrm2026} report $11.83$ points of
pass-at-$K$ improvement but only $1.37$ points with halting-head selection;
\citet{corbett2026} report guided accuracy $85.3\%$, below the $86.6\%$
deterministic baseline despite a $90.9\%$ oracle.
Our contribution is complementary paired accounting and site comparisons.

For HRM/Sudoku ($N=4096$, $k=8$), we retain each candidate's final q-halt
logit and compare within-puzzle spread between baseline errors for which
some rerun is correct and errors for which none is correct:

\begin{center}\small
\begin{tabular}{@{}lccc@{}}
\toprule
site & rescuable wrong & stuck wrong & ratio \\
     & (median q-halt SD) & (median SD) & \\
\midrule
$\zH$   & $6.22$ \, ($n{=}191$) & $0.42$ \, ($n{=}2202$) & $14.8\times$ \\
$\zL$   & $6.05$ \, ($n{=}121$) & $0.32$ \, ($n{=}2272$) & $18.9\times$ \\
input   & $5.65$ \, ($n{=}279$) & $1.48$ \, ($n{=}2114$) & $3.8\times$ \\
\bottomrule
\end{tabular}
\end{center}

\noindent Recurrent-site logit spread is larger for recoverable errors.
Input injection produces more spread overall but a smaller contrast between
the groups and worse selected accuracy (App.~\ref{app:ablation}).
These associations support examining guide alignment alongside candidate
diversity; spread alone does not establish correct ranking.
The corresponding rescuable-versus-unrecoverable comparison has not been
measured on Maze or TRM.

\section{TRM replication with shared modules}
\label{app:trm}
\begin{figure*}[t]
\centering
\includegraphics[width=\linewidth]{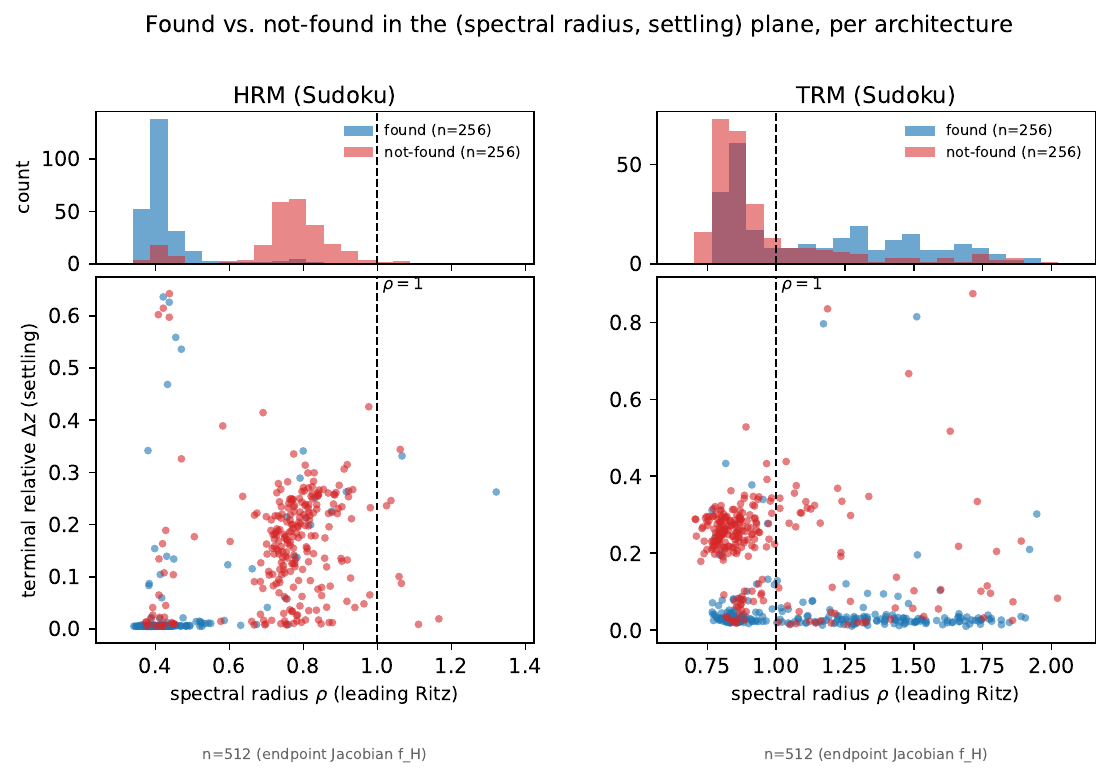}
\caption{\textbf{The inversion, per instance.} Every evaluated puzzle placed by
its endpoint Jacobian spectral radius $\rho$ (leading Ritz, $x$) and terminal
relative update (settling, $y$), coloured by whether the model found the
solution; top strips are the $\rho$ histograms. \emph{HRM} (left): found answers
contract to a tight $\rho\!\approx\!0.4$, low-residual cluster and failures spread
toward $\rho{=}1$. \emph{TRM} (right): the ordering \emph{inverts}---found answers
sit at $\rho{>}1$ (marginally expanding) yet still clear the terminal-update gate
(low residual), while failures fall to \emph{lower} $\rho$; the metastable
high-residual band (red) survives in both. That drop is carried by the high-residual band: gated to states
that pass the terminal-update gate, the two TRM strata coincide (see text).
Tables~\ref{tab:trm-split}--\ref{tab:trm-geom} made per-instance. ($n{=}512$ per
architecture.)}
\label{fig:found-distribution}
\end{figure*}

The pilot in App.~\ref{app:gram} changes the architecture but keeps two
\emph{separate} modules. This appendix removes the degree of freedom the main
text leans on hardest---the two modules themselves---at full protocol strength. TRM \citep{trm2025} is the strongest form of
Sec.~\ref{sec:discussion}'s question---\emph{one shared network} $f$ drives both
states on the same nested clock structure as HRM, $\zL\!\leftarrow\!f(\zL,\zH{+}e_x)$ run $L{=}6$
times then $\zH\!\leftarrow\!f(\zH,\zL)$ once, for $H{=}3$ high-level cycles per
segment over $16$ segments. We train it to $87.31\%$ exact on Sudoku-Extreme
(matching the paper's $\sim\!87.4\%$). Its own published ablation is independent
training-side evidence for the reading here: replacing the $1$-step gradient with
backpropagation through the whole final high-level cycle, its $\zL$ updates
together with the $\zH$ update that follows them, lifts accuracy $56.5\to87.4$.
The earlier high-level cycles still run under \texttt{no\_grad} and the carry is
detached between segments, so the comparison shows that non-terminal updates
carry learned computation inside one cycle rather than locating it across the
whole trajectory. We re-run all three headline analyses
verbatim on the EMA checkpoint, with the instrumented loop bit-exact to the
model's own forward. Figure~\ref{fig:found-distribution}
previews the headline: the found/not-found spectral split is the mirror image of
HRM's.

\paragraph{On unsolved states the timescale split inverts (Table~\ref{tab:trm-split}).}
Under the same held-out full-clock refit as Sec.~\ref{sec:transient}, the
\emph{fast} $\zL$ loop is the \emph{less linearisable} stream (unsolved
$R^2{=}0.45$, $n{=}48$ reliable high-energy buckets, split-half
$R^2_{\mathrm{corr}}{=}0.97$) and the \emph{slow} $\zH$ loop the more linear
($0.62$, $n{=}2$, $0.99$)---the mirror image of HRM's unsolved fast $0.80\!\gg\!$ slow
$0.39$. The robust anchor is the well-powered fast-$\zL$ estimate ($0.45$, far
below HRM's unsolved $0.80$); the slow-loop number rests on the two reliable high-energy
buckets the slow stream admits---a structural readout limit stable across
independent samples, not a tuned cut. Sharing the module does not preserve the
fast-more-linearisable/slow-less-linearisable dissociation on unsolved states; it
\emph{inverts} it there, and ties on solved.

\begin{table}[h]
\centering \small
\caption{TRM held-out high-energy per-tick linear $R^2$ (full-clock refit) at two
sample sizes, with HRM for reference. In TRM the fast $\zL$ loop is the
\emph{less linearisable} stream on the unsolved stratum---the inversion---and that
survives a $4\times$ larger sample. $n$ is the count of reliable high-energy
buckets surviving the $0.3$ reliability-ceiling cut; the retained count need not grow with sample size.}
\label{tab:trm-split}
\fittab{%
\begin{tabular}{lcccc}
\toprule
 & \multicolumn{2}{c}{$N{=}1024$} & \multicolumn{2}{c}{$N{=}4096$} \\
\cmidrule(lr){2-3}\cmidrule(lr){4-5}
 & fast $\zL$ & slow $\zH$ & fast $\zL$ & slow $\zH$ \\
\midrule
TRM, solved   & $0.43$ \, ($92$) & $0.35$ \, ($13$) & $0.48$ \, ($87$) & $0.48$ \, ($3$) \\
TRM, unsolved & $0.45$ \, ($48$) & $0.62$ \, ($2$)  & $0.45$ \, ($48$) & $0.60$ \, ($2$) \\
HRM, solved   & $\mathbf{0.72}$  & $0.37$ &---&---\\
HRM, unsolved & $0.80$           & $0.39$ &---&---\\
\bottomrule
\end{tabular}}
\end{table}

\paragraph{Reliable-tick coverage.}
Increasing the sample from $1024$ to $4096$ leaves the unsolved slow-loop
estimate at two reliable high-energy ticks and changes its $R^2$ from
$0.62$ to $0.60$. The solved slow-loop count changes from $13$ to $3$,
and its initial fast--slow gap disappears ($0.48/0.48$ at $N=4096$).
The reported inversion is therefore restricted to the unsolved stratum.

Tick count is bounded by the clock ($47$ slow versus $287$ fast ticks),
but the number passing the reliability floor can depend on sample size and
thresholds. The larger run does not establish that no further sample could
improve coverage. HRM's default slow-loop aggregate also uses only $1$--$3$
ticks; App.~\ref{app:sensitivity} reports how relaxing the thresholds
increases coverage. Conclusions concern the ordering under the stated
estimator, with limited slow-tick coverage.

\paragraph{Persistence and raw spectra.}
TRM answer persistence is $0.96$ solved versus $0.32$ wrong, and state
persistence is $0.32/0.02$ (Table~\ref{tab:trm-geom}).
The wrong class pools all three severity buckets ($96$ endpoints per class);
the HRM pilot uses only mild failures ($32$ per class). Thus the protocols
share radii and criteria, but their absolute rates are not a severity-matched
comparison. A mild-only TRM rate cannot be inferred from HRM's severity pattern.

Both original sweeps omit $\varepsilon=0$ and retain only radius aggregates,
so endpoint-clustered intervals cannot be recovered from those artifacts.
Separate unperturbed controls are reported in App.~\ref{app:eps0}.
The raw TRM leading-Ritz means are $1.15$ solved and $0.96$ wrong
(wrong maximum $2.02$), with correlation $-0.42$ between radius and gate
failure. These local estimates do not establish attractor stability.

\paragraph{The residual gate dissolves the inversion.}
Only $24\%$ of the $256$ wrong TRM endpoints pass the terminal-update
gate. Applying Eq.~\ref{eq:gate} with TRM's own solved calibration
($q_{95}=0.148$) gives mean radii $1.149$ for wrong endpoints ($n=62$)
and $1.151$ for solved endpoints ($n=243$). The reported unrounded
difference is $0.003$ ($95\%$ bootstrap CI $[-0.09,0.09]$;
Mann--Whitney $p=0.57$).
The excluded $194$ wrong endpoints have mean radius $0.90$ and account
for the raw reversal. HRM retains a gated difference ($0.43$ solved,
$n=243$, versus $0.73$ wrong, $n=176$; $p<10^{-40}$).
Thus no spectral ordering is detected in gate-passing TRM endpoints.
Because the gate is calibrated to solved residuals, passing it does not
certify that an endpoint is a root.

\begin{table}[h]
\centering \small
\caption{TRM endpoint geometry under Sec.~\ref{sec:fail}'s stability test, with
HRM anchors. Every cell is a class \emph{mean} ($\bar\rho$, gate-passing) or the
median over the $13$-radius sweep ($R_{\mathrm{ans}}$, $R_{\mathrm{state}}$); no
column mixes a mean with a max. The solved-vs-wrong ordering agrees between the
two models in every column but $\bar\rho$, whose sign flips (and vanishes under
the gate); $R_{\mathrm{ans}}$ admits the better-separated
threshold for these two settings: $\tfrac12$ leaves all four cells at least
$0.12$ from the cut, while a state-level cut of $0.22$ separates the same four
cells with a margin of $0.10$. Neither cut travels beyond these two settings
(Fig.~\ref{fig:coverage}). The $\bar\rho$ column is \emph{ungated} and averages
gate-passing and gate-failing endpoints together; after gating the TRM rows
coincide ($1.151$ vs $1.149$) and only the gate-passing fraction still separates
them. Severity coverage of the two wrong rows differs (see text).}
\label{tab:trm-geom}
\begin{tabular}{lcccc}
\toprule
 & $\bar\rho$ & gate-passing & $R_{\mathrm{ans}}$ & $R_{\mathrm{state}}$ \\
\midrule
TRM, solved & $1.15$ & $0.95$ & $0.96$ & $0.32$ \\
TRM, wrong  & $0.96$ & $0.24$ & $0.32$ & $0.02$ \\
HRM, solved & $0.44$ & $0.95$ & $0.97$ & $0.81$ \\
HRM, wrong  & $0.75$ & $0.69$ & $0.38$ & $0.12$ \\
\bottomrule
\end{tabular}
\end{table}

\paragraph{Reselection at both TRM sites (Table~\ref{tab:trm-interv}).}
On $N=2048$ puzzles, injecting $z_L$ gains $8.25$ points
($95\%$ CI $[7.1,9.4]$, $169/0$); injecting $z_H$ gains $8.40$
($172/0$). Their baselines differ slightly ($0.852$ and $0.849$), and
we do not have a paired between-site test establishing equivalence.
The observed gains provide no clear evidence that the less-linearisable
stream is the preferred site. Perturbation and halting-head selection
are established in \citet{ptrm2026}, which reports $87.28\%\to98.75\%$
at a larger budget; these are different protocols and checkpoints.

\begin{table}[h]
\centering \small
\caption{Oracle-free noise-injection $+$ halting-logit reselection, TRM (inject
the fast $\zL$ loop) vs HRM (inject the slow $\zH$ loop). Each injects into the
loop its own full-clock refit finds least linearisable. In TRM the two loops are
indistinguishable as sites ($+8.25$ vs.\ $+8.40$), so that choice is not what
carries the gain there.}
\label{tab:trm-interv}
\fittab{%
\begin{tabular}{lcccc}
\toprule
 & site & exact & $\Delta$ (95\% CI) & gained/lost \\
\midrule
TRM & $\zL$ & $.852{\to}.934$ & $+.083\,[{+}.071,{+}.094]$ & $169/0$ \\
TRM & $\zH$ & $.849{\to}.933$ & $+.084\,[{+}.072,{+}.096]$ & $172/0$ \\
HRM & $\zH$ &---            & $+.048\,[{+}.041,{+}.054]$ & $197/2$ \\
\bottomrule
\end{tabular}}
\end{table}

\paragraph{Additional coverage qualifications.}
N-Queens-$10\times10$ has answer-persistence rates $0.96/0.56$ for
solved/wrong endpoints. Valid but non-canonical boards give $0.93$,
close to the solved rate; they should not be interpreted as constraint
violations. Input injection gives the largest observed gain on this task
($8.8$ points), so the preferred HRM injection site does not transfer.
The N-Queens and GRAM-guided sweeps omit $\varepsilon=0$; separate unperturbed
controls on the same checkpoints return answer persistence $0.97/0.59$ and
$0.98/0.84$. Each control is one deterministic continuation per endpoint
($64$ per class), not the $768$ continuations behind the perturbed cells.

The Maze full-clock and intervention cells and the seed-1 full-clock cell are
stored metrics, not values quoted from manuscript tables: the seed-1 refit
records $11$ reliable fast buckets against $1$ slow bucket, and the Maze
epoch-$7000$ refit $18$ against $3$. The third-party GRAM endpoint pilot and the
GRAM-guided TRM replication are distinct experiments
(App.~\ref{app:gram}); their results must not be used interchangeably.

\paragraph{Cross-architecture scope.}
Solved--wrong persistence ordering and positive reselection gains appear in
all six settings, but state-persistence thresholds use setting-specific
scales ($\kappa=1.4$--$47$, across $81$--$900$ positions). The answer-level
$\tfrac12$ cut also fails on both N-Queens settings, despite their newly
measured unperturbed controls. Endpoint and full-clock patterns transfer less
consistently. Both TRM variants differ from HRM in several respects, so module
sharing is not isolated as a cause; GRAM-guided TRM also lacks HRM's
terminal-gate separation ($94.9\%$ solved / $96.5\%$ wrong). Finally, TRM's
two injection sites yield similar gains despite different linearity estimates.
The transferable observation is the persistence ordering, not a universal
spectral, residual or optimal-injection-site rule.

\end{document}